%% file: neurips_2026.tex
\documentclass{article}

\usepackage[preprint]{neurips_2026}

\usepackage[utf8]{inputenc} % allow utf-8 input
\usepackage[T1]{fontenc}    % use 8-bit T1 fonts
\usepackage{hyperref}       % hyperlinks
\usepackage{url}            % simple URL typesetting
\usepackage{booktabs}       % professional-quality tables
\usepackage{amsfonts}       % blackboard math symbols
\usepackage{nicefrac}       % compact symbols for 1/2, etc.
\usepackage{microtype}      % microtypography
\usepackage{xcolor}         % colors

\usepackage{xcolor}
\usepackage{float}
\usepackage{afterpage}
\usepackage{pifont} 
\newcommand{\cmark}{\textcolor{green}{\ding{51}}}  % Check mark
\newcommand{\xmark}{\textcolor{red}{\ding{55}}}    % Cross mark
\newcommand{\tabitem}{\hspace{1em}\textbullet\hspace{0.5em}}

\usepackage{caption}
\usepackage{makecell}
\usepackage{tabularx}
\usepackage{colortbl}
\usepackage{amsmath}
\usepackage{graphicx}
\usepackage{enumitem}
\usepackage{multicol}
\usepackage{multirow}
\usepackage{wrapfig,lipsum,booktabs}
\usepackage[ruled]{algorithm2e} % For 

\SetAlFnt{\small}
\SetAlCapFnt{\small}
\SetAlCapNameFnt{\small}
\SetAlCapHSkip{0pt}

\newcommand\blfootnote[1]{%
  \begingroup
  \renewcommand\thefootnote{}\footnote{#1}%
  \addtocounter{footnote}{-1}%
  \endgroup
}

\title{HOMIE: \underline{H}uman-\underline{o}bject Centric Video Personalization via \underline{M}ultimodal \underline{I}ntelligent \underline{E}nhancement}

\author{%
  \textbf{Yiyang Cai}\textsuperscript{*}, \textbf{Nan Chen}\textsuperscript{*}, \textbf{Rongchang Xie}, \textbf{Junwen Pan}, \textbf{Chunyang Jiang}, \textbf{Cheng Chen}, \\
  \textbf{Wen Zhou}, \textbf{Zhenbang Sun}, \textbf{Wei Xue}, \textbf{Wenhan Luo}, \textbf{Yike Guo} \\
  Hong Kong University of Science and Technology \\
  Page: \url{https://yiyangcai.github.io/homie-page.github.io/}
}

\begin{document}

\maketitle
% \blfootnote{$^\dagger$ Corresponding Authors.}
% \blfootnote{$\ddagger$ Project Leader.}
\blfootnote{\textsuperscript{*} Equal Contribution.}

\begin{center}
    \vspace{-20pt}
    \captionsetup{type=figure}
    \includegraphics[width=0.99\textwidth]{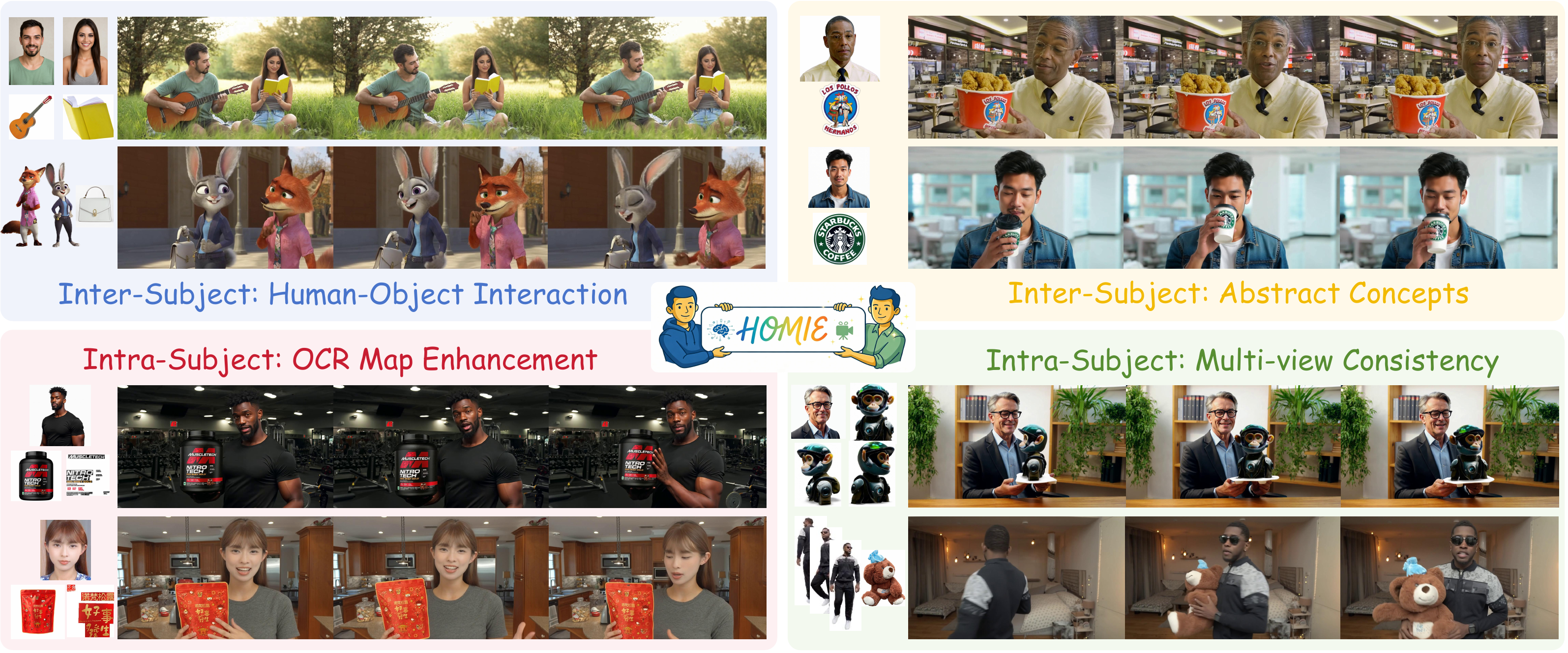}
    \caption{HOMIE addresses HOCVP with both inter/intra-subject references: (1) multi-human-object personalization with diverse interaction patterns; (2) abstract concept personalization, which automatically links abstract references to the most relevant objects in the video without explicitly mentioning such a relationship in the prompt; and (3) intra-subject personalization. For the scenario of intra-subject, by processing multiple references of the same subject, HOMIE enables targeted enhancements, such as leveraging OCR maps for improved textual fidelity and utilizing multi-view inputs to increase spatial consistency during interactions (\textit{Zoom in for better view}).}
    \label{fig:teaser}
\end{center}

\begin{abstract}
Human-object centric video personalization (HOCVP) is a core task within subject-driven video generation. However, existing methods suffer from two key limitations. First, most approaches focusing on \textit{inter-subject} personalization still struggle to strike a balance between high subject fidelity and accurate interaction patterns between humans and diverse objects, especially when objects represent abstract concepts such as logos. Second, while \textit{intra-subject} references (\textit{e.g.}, OCR maps, multi-view inputs) are expected to enhance subject fidelity, most existing works lack mechanisms to understand such latent correspondence.
To address both challenges, we propose HOMIE, an HOCVP framework that tackles both inter- and intra-subject input settings in a unified manner. Compared to previous approaches, HOMIE proposes a better MLLM integration strategy to extract knowledge of reference-level relationships without compromising the controllability of text encoders or incurring costly re-alignment. Specifically, we introduce global multimodal guidance within self-attention to better align MLLM-derived semantic features with VAE tokens. Furthermore, we propose modality-reference embedding to differentiate tokens from MLLM features and VAE tokens and associate intra-subject reference image tokens. Extensive experiments validate that our method achieves state-of-the-art performance across various HOCVP tasks.
\end{abstract}

\input{1_introduction}

\input{2_related_work}

\input{3_method}

\input{4_experiments}

\input{5_conclusion}

% =============== Original Template Related Contents (to be removed ) ===================
% \section*{References}
\bibliographystyle{plain}
\bibliography{main}

% References follow the acknowledgments in the camera-ready paper. Use unnumbered first-level heading for
% the references. Any choice of citation style is acceptable as long as you are
% consistent. It is permissible to reduce the font size to \verb+small+ (9 point)
% when listing the references.
% Note that the Reference section does not count towards the page limit.
% \medskip

% {
% \small

% [1] Alexander, J.A.\ \& Mozer, M.C.\ (1995) Template-based algorithms for
% connectionist rule extraction. In G.\ Tesauro, D.S.\ Touretzky and T.K.\ Leen
% (eds.), {\it Advances in Neural Information Processing Systems 7},
% pp.\ 609--616. Cambridge, MA: MIT Press.

% [2] Bower, J.M.\ \& Beeman, D.\ (1995) {\it The Book of GENESIS: Exploring
%   Realistic Neural Models with the GEneral NEural SImulation System.}  New York:
% TELOS/Springer--Verlag.

% [3] Hasselmo, M.E., Schnell, E.\ \& Barkai, E.\ (1995) Dynamics of learning and
% recall at excitatory recurrent synapses and cholinergic modulation in rat
% hippocampal region CA3. {\it Journal of Neuroscience} {\bf 15}(7):5249-5262.
% }

%%%%%%%%%%%%%%%%%%%%%%%%%%%%%%%%%%%%%%%%%%%%%%%%%%%%%%%%%%%%
\newpage
\appendix
\input{x_supp}

%%%%%%%%%%%%%%%%%%%%%%%%%%%%%%%%%%%%%%%%%%%%%%%%%%%%%%%%%%%%

% \clearpage
% \input{checklist.tex}

\end{document}

%% file: 1_introduction.tex
\section{Introduction}
\label{sec:intro}

% \begin{figure*}
%     \centering
%   \includegraphics[width=1.0\textwidth]{images/1_teaser.pdf}
%   \caption{HOMIE teaser text here. (\textbf{Zoom in for the best view})}
%   \label{fig:teaser}
% \end{figure*}

% 第一段：介绍背景应用，任务定义和一些简要的文献
Human-object centric video personalization (HOCVP) stands out as a high-value and practical research direction in controllable video generation. By taking reference images of humans and objects as inputs, HOCVP synthesizes videos featuring diverse human-object interactions, unlocking broad application scenarios and immense commercial potential. To advance this field, substantial efforts have been dedicated to both methodological innovations \cite{phantom, deng2026magref} and dataset curation \cite{chen2026phantomdata, yuan2026opensvnexus}. These developments have enabled HOCVP to achieve superior subject consistency and controllability.

Despite recent progress, existing HOCVP methods face several challenges. We categorize HOCVP scenarios into two types based on reference structure (see Fig. \ref{fig:teaser}): \textbf{inter-subject HOCVP}, where all reference images depict distinct subjects, and \textbf{intra-subject HOCVP}, where multiple references correspond to the same subject (\textit{e.g.}, multi-view images or OCR maps of a single object, aiming at improving multi-view consistency or text fidelity, respectively). In inter-subject settings, as the number of subjects increases, current methods struggle to balance fidelity with precise interactions, often yielding copy-paste artifacts and restricted expressiveness. Furthermore, they lack the inferential reasoning needed to handle abstract references such as logos, making it difficult to place them accurately within generated videos. In intra-subject settings, existing methods often lack effective mechanisms for extracting semantic correspondences, or else depend on supervision from 3D multi-view priors. Treating these components in isolation tends to produce redundant outputs and results in limited fidelity improvements.

\begin{figure*}[h]
\includegraphics[width=1\linewidth]{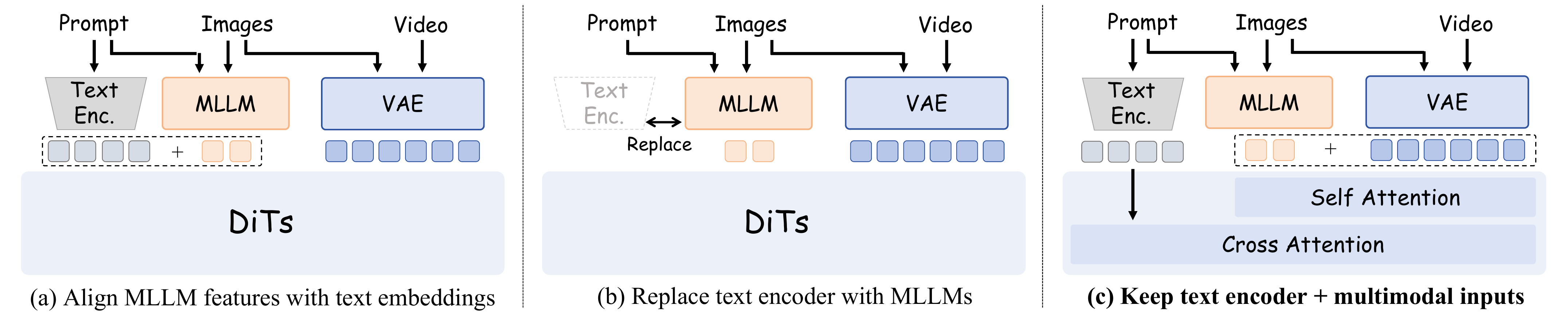}
\vspace{-5pt}
\caption{Framework design of MLLM-facilitated video personalization.}
\label{fig:mllm_design}
\vspace{-5pt}
\end{figure*}

% 第三段的最新逻辑（梳理MLLM-based 方法）
Resolving both types of challenges requires a model capable of reasoning implicit associations among references, motivating the integration of Multimodal Large Language Models (MLLMs) into video diffusion models.
While MLLM integration has proven effective in image generation \cite{step1x-edit, uniworld, metaquery, wu2025omnigen2}, current approaches that integrate MLLMs into HOCVP tasks still struggle to address the aforementioned challenges due to their integration strategies.
Currently, these approaches follow two main paths. First, some methods \cite{li2026bindweave, fei2025skyreels} \textbf{align} MLLM features with the UmT5 textual embedding space in cross-attention-based diffusion models \cite{lin2024open, hacohen2024ltx, polyak2024movie, wan2025wan} (Fig.~\ref{fig:mllm_design}(a)). This diminishes UmT5's established controllability and bottlenecks MLLM features within a constrained space, thereby restricting its ability to infer precise visual inter-subject associations (\textit{e.g.}, logo placements) as well as intra-subject relationships. Second, other frameworks \textbf{replace} the text encoder entirely with an MLLM, connecting its outputs to the DiT via hidden states or learnable queries \cite{wei2026univideo, mou2025instructx, chen2026vino, lin2025exploring, pan2026omniweaving} (Fig.~\ref{fig:mllm_design}(b)). Such a strategy incurs substantial re-alignment costs and shoulders the MLLM with a heavy burden of reconstructing general text-to-video control, distracting it from prioritizing knowledge extraction of inter- and intra-subject reference relationships. To overcome these limitations, we propose the paradigm (Fig.~\ref{fig:mllm_design}(c)) that preserves the text encoder, enabling the MLLM to focus exclusively on extracting reference relationships. By constructing a unified multimodal input stream of video, reference images, and MLLM tokens, this architecture facilitates robust cross-modal interaction without requiring costly realignment.

Building on the integration strategy discussed above, we introduce HOMIE, an HOCVP framework that more effectively leverages MLLM knowledge of both inter- and intra-subject relationships. By processing unified multimodal inputs, HOMIE employs two key components to improve generation performance. First, it incorporates \textbf{Global Multimodal Guidance (GMG)}, which derives global representations from MLLMs during the query-key computation stage. These representations enrich the queries and keys of video tokens with temporal interaction information, thereby seamlessly injecting MLLM knowledge into the self-attention process. In addition, HOMIE introduces a simple yet effective module called \textbf{Modality-Reference Embedding (MRE)}. This module explicitly distinguishes tokens by their input modalities while jointly associating multiple intra-subject references in a unified representation space. Through these designs, HOMIE consistently enhances HOCVP performance across multiple metrics, including prompt adherence and subject consistency, under diverse inter- and intra-subject input settings.

% In summary, our principal contributions are threefold:
% summary here

In summary, our contributions are threefold:
\begin{itemize}[topsep=0pt,leftmargin=*]
  \item We propose HOMIE, a unified framework for both inter- and intra-subject HOCVP, enhanced by knowledge from MLLMs. We introduce Global Multimodal Guidance (GMG), which injects multimodal features into video tokens to improve semantic reasoning and interaction modeling.
  \item We design Modality-Reference Embedding (MRE), which distinguishes features from the same modality. It also captures intra-subject associations and inter-subject distinctions, thereby improving the identity fidelity of intra-subject HOCVP.
  \item HOMIE demonstrates competitive performance. Extensive evaluations validate HOMIE's robustness across diverse inter- and intra-subject HOCVP scenarios, including general inter-subject personalization and intra-subject personalization with OCR maps and multi-view images.
\end{itemize}

%% file: 2_related_work.tex
\section{Related Work}
\label{related_work}
\textbf{Video Diffusion Models}.
The success of text-to-image diffusion models \cite{latent_diffusion_model, bflflux} has catalyzed efforts to extend these techniques into the video domain. Recently, video diffusion models (VDMs) have advanced significantly through architectural innovations and high-quality data curation. These models generally operate within text-to-video (T2V) or image-to-video (I2V) paradigms. Architecturally, early U-Net-based frameworks \cite{ho2022video, svd} are increasingly superseded by state-of-the-art diffusion transformers (DiTs) \cite{wan2025wan, yang2024cogvideox, kong2024hunyuanvideo, seedance2025seedance}, yielding superior visual fidelity and temporal coherence. This rapid evolution provides a robust foundation for diverse downstream applications, including video editing \cite{qi2023fatezero, vace, ye2026unified}, digital human animation \cite{guo2024animatediff, kong2026let, dreamactor, Anchorcrafter, tong2026mvhoi, zhong2025anytalker}, and personalized video synthesis \cite{wei2024dreamvideo, ConsisID, VideoAlchemist, 10.1145/3746027.3761989, abdal2025dynamic, 10.1145/3757377.3763949, xue2025stand, mai2025contextanyone, caiomnivcus, ling2026mofu, xing2026lumosx, guo2026wildactor, chen2026domainshuttle}.

\noindent\textbf{Video Personalization}.
Inspired by the concept of image personalization \cite{ruiz2023dreambooth, guo2024pulid, cai2026foundation}, video personalization generates subject-consistent videos from reference images by retaining fine-grained target features. While early studies \cite{ID-Animator, ConsisID, PersonalVideo, MagicID, xue2025stand} focus on single-identity generation, recent efforts target multi-subject personalization—a challenging but practical task highly aligned with HOCVP. To handle multiple subjects, \cite{ConceptMaster, VideoAlchemist, sang2025lynx} employ decoupled cross-attention to mitigate attribute mixing, while MovieWeaver \cite{MovieWeaver} binds identity features to specific tokens via prompt anchors. Additionally, Phantom \cite{phantom} dynamically injects references to handle varying counts, and methods like MAGREF and FFGO \cite{deng2026magref, chen2025first} leverage image-to-video animation capabilities by embedding references into the first frame. Some works \cite{wu2026consid, songmv} use supervision derived from 3D priors to strengthen multi-view consistency for intra-subject inputs. Recently, several works \cite{deng2025cinema, hunyuancustom, fei2025skyreels, li2026bindweave, pan2025id, hu2025polyvivid} have integrated MLLMs \cite{qwen25, llava} to enhance controllability. RefAlign \cite{wang2026refalign} tries to disentangle inter-subject references via the feedback from MLLMs. However, these approaches face inherent limitations in their multimodal feature integration strategies; furthermore, most of them focus on inter-subject scenarios. These limitations jointly hinder the full utilization of MLLM knowledge for improving overall generation quality over diverse HOCVP scenarios.

%% file: 3_method.tex
\section{Method}
\label{method}

\begin{figure*}[t]
    \includegraphics[width=1.00\linewidth]{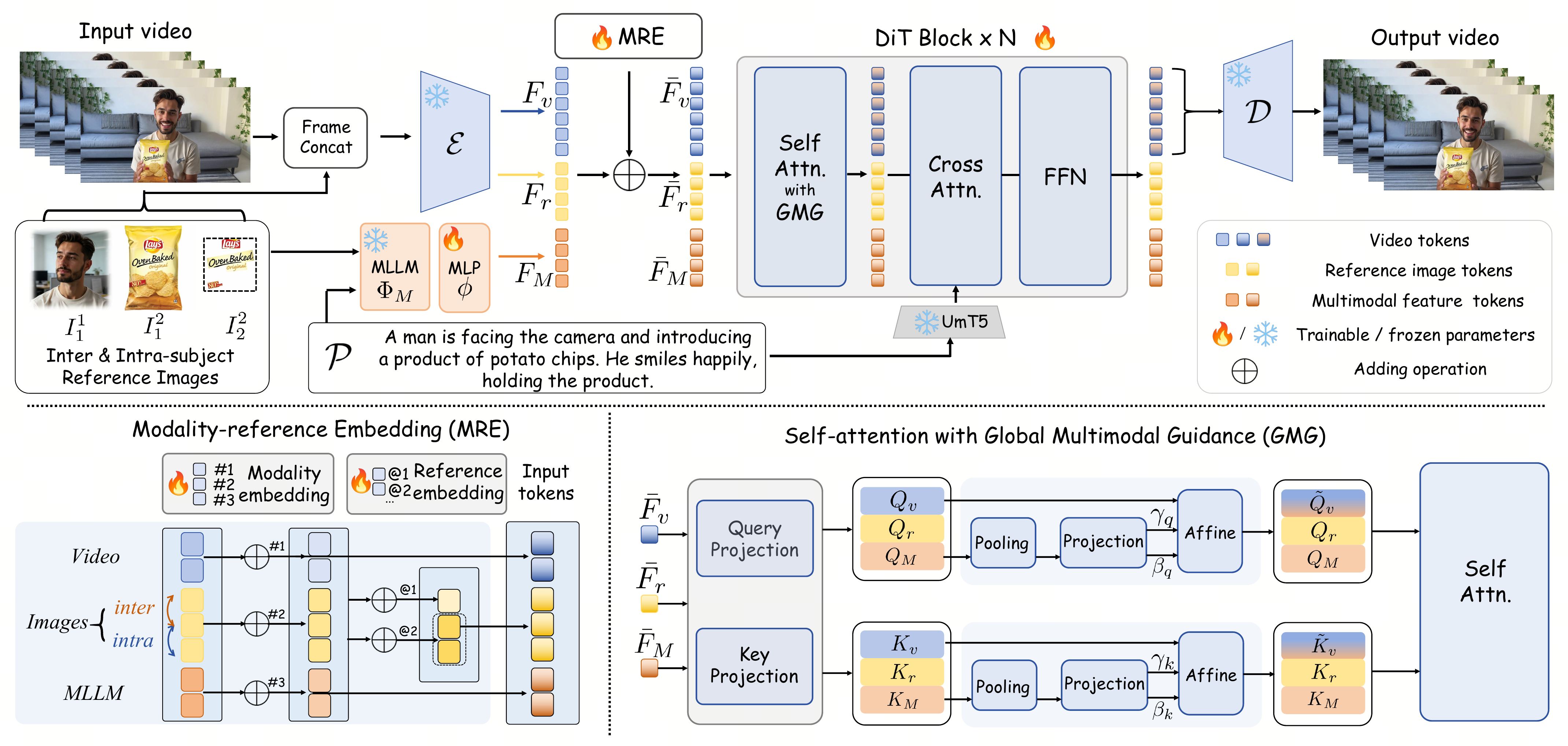}
    \caption{\textbf{Overview of HOMIE.} HOMIE employs a multimodal-input paradigm with video tokens, image tokens, and multimodal features. Prior to self-attention, Global Multimodal Guidance (GMG) fuses high-semantic MLLM features into video tokens, boosting cross-modal interaction. Furthermore, Multimodal-Reference Embedding (MRE) enables the model to discriminate cross-modal tokens and inter-subject tokens, as well as bind reference tokens from intra-subject references.}
    \label{fig:method_overview}
\end{figure*}

% \subsection{Preliminaries} % delete this title
% ============= refined ==============
We develop our method based on a DiT-based text-to-video model \cite{wan2025wan} as our backbone architecture. As illustrated in Fig. \ref{fig:method_overview}, HOMIE integrates a 3D Variational Autoencoder (3D VAE) to compress raw video into a compact latent space. A text encoder \cite{chung2023unimax} is employed to encode textual prompts, which are then injected into the DiT backbone via cross-attention mechanisms, establishing robust text-video semantic alignment. For model training, it leverages the flow matching paradigm \cite{lipman2023flow}, which learns a continuous velocity field to transform samples from a simple prior distribution to the target data distribution along deterministic trajectories. The training objective $\mathcal{L}_{\mathrm{FM}}(\theta)$ is:
\begin{equation}
\mathcal{L}_{\mathrm{FM}}(\theta)=\mathbb{E}_{t,z_0,z_1}\|{v}_\theta(z_t,t,c)-(z_1-z_0)\|_2^2,
\label{eq:flow_matching_loss}
\end{equation}
where $z_0$ denotes a sample drawn from a prior distribution, $z_1$ is the latent of the target video sample, and $c$ denotes the condition (set). The DiT model $v_\theta$ takes the noisy latent $z_t = (1 - t){z}_0 + tz_1$, and learns to predict its velocity at any point $t\in[0,1]$. In our proposed HOMIE framework, we extend the conditioning set $c$ as $c= \{{c}_{\text{txt}},c_{img}\}$ to incorporate multimodal data inputs, including reference images and multimodal features to enable high-fidelity and controllable HOCVP.

\subsection{HOMIE}
% ============= refined ==============
HOMIE is designed to advance HOCVP, where both inter- and intra-subject references and corresponding complex interaction patterns are included via MLLM integration. In Sec.\ref{sec:input_paradigm}, we first outline our multimodal input paradigm, which consists of raw videos, reference images, and textual conditions. Reference images and prompts are fed into an MLLM to obtain highly semantic features. We then present \textbf{Global Multimodal Guidance (GMG)} in Sec.\ref{sec:gmg}, a dedicated fusion mechanism integrated into self-attention layers that effectively injects multimodal features into the video latents. Next, we introduce \textbf{Modality-Reference Embedding (MRE)} in Sec.\ref{sec:mre}, a learnable embedding module that discriminates between distinct modality types of input tokens and identifies inter-/intra-subject reference tokens. Finally, we introduce the dataset curation pipeline, which enables the deployment of a multi-stage training strategy to gradually optimize model performance. The complete HOMIE pipeline is illustrated in Fig.\ref{fig:method_overview}.

\subsubsection{Multimodal Input Paradigm with Inter- and Intra-subject References}
\label{sec:input_paradigm}
% ============= refined ==============

HOMIE accepts a textual prompt $\mathcal{P}$ and a set of reference images $\mathcal{I} = \{I^m_n\}$ as inputs. Here, superscript $m$ denotes subject identity (identical $m$ indicates intra-subject references), and subscript $n$ indexes the image instance. Following standard architectural paradigms, $\mathcal{I}$ is mapped into a latent space via a pre-trained 3D VAE. Concurrently, to capture the complex semantic alignment between visual subjects and textual instructions, an MLLM ($\Phi_{M}$) processes both $\mathcal{I}$ and $\mathcal{P}$ to extract a multimodal feature $F_{M}$. To align this conditioning feature $F_{M}$ with the 3D-VAE token space, we introduce a lightweight alignment network $\phi$:

\begin{equation}
F_{M} = \phi(\Phi_{M}(\mathcal{P}_{sys}, \mathcal{P}, \mathcal{I})),
\end{equation}

where $\mathcal{P}_{sys}$ denotes the system prompt of $\Phi_{M}$. Unlike prior methods \cite{li2026bindweave, fei2025skyreels} that concatenate multimodal features to UmT5 embeddings, we posit that $F_M$ contains richer visual details and linguistic context. Therefore, we integrate it directly into the generation process by concatenating it with the latent tokens of the target video, denoted as $F_{v}$, and reference images, denoted as $F_{r}$:
\begin{equation}
F = [F_{v} \oplus F_{r} \oplus F_{M}],
\end{equation}
where $\oplus$ denotes the concatenation operation. Such a combination on the input side will allow more intrinsic interaction of tokens from different modalities. 

\subsubsection{Global Multimodal Guidance}
\label{sec:gmg}

However, because $F_{M}$ retains highly semantic information that may still exhibit a distributional gap relative to the 3D VAE features, standard self-attention mechanisms risk disrupting the model's intrinsic feature interactions. Furthermore, this multimodal feature must demonstrate stable control across the temporal dimension. Therefore, we introduce a simple yet effective strategy termed \textbf{Global Multimodal Guidance (GMG)}, which explicitly injects semantic information from $F_{M}$ into $F_{v}$. Its mechanism is illustrated in the bottom-right portion of Fig.\ref{fig:method_overview}. After projecting the query and key matrices, we partition them into components corresponding to the video tokens, reference tokens, and MLLM feature tokens. Denoting the partitioned query and key matrices for the video, reference, and multimodal features as $\{Q_{v}, Q_{r}, Q_{M}\}$ and $\{K_{v}, K_{r}, K_{M}\}$ respectively, we restrict our enhancement strictly to the interaction between the video latents and the MLLM features. The query and key matrices of the reference images $\{Q_{r},K_{r}\}$ remain unchanged to preserve their native self-attention patterns with the video tokens. Specifically, considering the matrices $Q_{v} \in \mathbf{R}^{b\times l_{v} \times h}$ and $Q_{M} \in \mathbf{R}^{b\times l_{M} \times h}$, where $l_v$ and $l_M$ denote the sequence lengths of the video tokens and MLLM features respectively, we compute the global representations for the MLLM query ${Q}_{M}$ and key ${K}_{M}$ by applying a pooling operation along the temporal dimension:

\begin{equation}
    \tilde{Q}_{M} = \mbox{Pooling}(Q_{M}, \mbox{dim}=1) \in \mathbf{R}^{b\times 1 \times h},
\end{equation}
\begin{equation}
    \tilde{K}_{M} = \mbox{Pooling}(K_{M}, \mbox{dim}=1) \in \mathbf{R}^{b\times 1 \times h}.
\end{equation}
Subsequently, to explicitly condition the attention mechanism, we feed this global representation into two lightweight projection networks, $\gamma$ and $\beta$, to derive scaling and shifting modulation factors. After broadcasting these parameters to align with the sequence length $l_{v}$ of the video tokens, we apply a feature-wise affine transformation to $Q_{v}$ and $K_v$. The complete modulation process is mathematically formulated as follows:
\begin{equation}
    \tilde{Q}_{v} = (1+\gamma_{q}(\tilde{Q}_{M}))\times Q_{v}+ \beta_{q}(\tilde{Q}_{M}),
\end{equation}
\begin{equation}
    \tilde{K}_{v} = (1+\gamma_{k}(\tilde{K}_{M}))\times K_{v}+ \beta_{k}(\tilde{K}_{M}).
\end{equation}
Finally, we concatenate the partitioned query and key components back into their original matrices to execute the standard attention computation. Through the proposed GMG mechanism, high-level semantic multimodal guidance is effectively infused into the video tokens.
%Such process can be demonstrated by the bottom-right part of 

\subsubsection{Modality-Reference Embedding}
\label{sec:mre}
% ============= refined version ==============
HOMIE establishes a unified multimodal input paradigm consisting of video, reference, and multimodal token embeddings. While this design enriches the model with fine-grained semantic information, it also creates the need to distinguish and align representations across different modalities. Moreover, because inter- and intra-subject references correspond to different target identities, the model must also capture their underlying associations. To address these challenges in a once-for-all manner, we propose \textbf{Modality-Reference Embedding (MRE)}, a learnable embedding module designed to simultaneously annotate modality-specific attributes and enforce identity-level consistency across correlated inputs. As illustrated in the bottom-left region of Fig.\ref{fig:method_overview}, MRE consists of two components: modality embeddings (\textbf{ME}) and reference embeddings (\textbf{RE}). Specifically, prior to projecting the tokens into the stacked DiT blocks, learnable modality embeddings are selectively integrated into each token according to its source modality (\textit{i.e.}, patchified video latents, encoded reference image latents, or aligned multimodal features). This explicit embedding mechanism systematically enables the model to disentangle and accurately process these heterogeneous input representations:
\begin{equation}
    \bar{F}_{*} = {F}_{*}+\text{ME}[Proj(*)],
\end{equation}
where $*\in{[v, r, M]}$ denotes different modality types, and $Proj$ denotes a predefined projection function that maps each modality type to distinct slice indices. Subsequently, reference embeddings are assigned to tokens derived from reference inputs. Specifically, the corresponding embedding vector is selectively added to reference image tokens based on their original reference entities:
\begin{equation}
    \bar{F}^{m}_{r} = \bar{F}^{m}_{r}+\text{RE}[m],
\end{equation}
where $m$ denotes the $m^{th}$ unique reference entity fed into the model. Let us consider the reference images setting $\mathcal{I} = \{I^m_n \}$ mentioned before: references with the same annotation $m$ should be assigned with the same reference embedding, since they are intra-subject references which share the same identity information. Conversely, inter-subject references are assigned distinct reference embeddings to explicitly denote their disparate identities. By facilitating robust disentanglement across diverse modalities and inter-subject references while simultaneously binding intra-subject samples, MRE aligns seamlessly with the structural requirements of our multimodal input paradigm.

\subsection{Datasets Construction}
\label{sec:dataset_curation}
We construct the training dataset through a multi-step pipeline based on two open-source subject-driven video generation datasets: OpenS2V-5M \cite{yuan2026opensvnexus} and PhantomData \cite{chen2026phantomdata}. First, we remove low-quality videos using generic quality metrics such as aesthetic scores. Next, leveraging the reference annotations provided by these datasets, including category labels and area proportions, we filter out subjects and corresponding video clips that are irrelevant to HOCVP, such as backgrounds and small objects. The resulting data are then divided into two video-consistency subsets: a single-subject dataset containing only humans or objects, with 300K samples, and a multi-subject dataset containing both humans and objects, with 80K samples. To further improve performance, we additionally curate a high-quality dataset of 20K HOC video clips with precise segmentation masks for both humans and objects, and incorporate it into the multi-subject subset. Moreover, this self-curated dataset includes the same cropped object across different frames, enabling the construction of intra-subject references for training MRE. Finally, from the expanded 100K-sample multi-subject dataset, we further select approximately 40K high-resolution (720P) samples.

%% file: 4_experiments.tex
% main results form
\begin{table*}[h]
\centering
\caption{\textbf{Quantitative comparison results of HOCVP}. The best scores are shown in \textbf{bold}, and the second-best are \underline{underlined}.}
\setlength\tabcolsep{10pt}
\resizebox{1\textwidth}{!}{
\begin{tabular}{lcccccccc}
\toprule
\multirow{2}{*}{Method} & \multicolumn{2}{c}{Video Quality}  & Text Following  & \multicolumn{4}{c}{Subject Consistency} \\
\cmidrule(lr){2-3} \cmidrule(lr){4-4} \cmidrule(lr){5-8} & AES$\uparrow$ & MS$\uparrow$ & GMEScore$\uparrow$ & Face-Sim$\uparrow$ & DINO-I$\uparrow$ & Obj-Sim $\uparrow$ & OCR Acc.$\uparrow$ \\ \midrule
Kling 1.6 \cite{kling}              & \textbf{0.558} & \textbf{0.997} & 0.531 & 0.678 & 0.492 & 0.764 & 0.272 \\ \midrule
VACE \cite{vace}                & 0.517 & 0.994 & 0.685 & 0.650 & \underline{0.539} & 0.828 & 0.371 \\
MAGREF \cite{deng2026magref}        & 0.499 & 0.971 & 0.689 & 0.589 & 0.429 & 0.754 & 0.135 \\
SkyReels-A2 \cite{fei2025skyreels}  & 0.493 & 0.946 & 0.652 & 0.599 & 0.524 & 0.803 & 0.288 \\
SkyReels-V3 \cite{li2026skyreels}   & 0.523 & 0.971 & 0.656 & 0.739 & 0.526 & 0.847 & 0.326 \\
Phantom \cite{phantom}              & 0.505 & 0.991 & 0.677 & \textbf{0.801} & 0.519 & 0.774 & 0.302 \\
HuMo \cite{chen2026human}            & 0.468 & 0.988 & 0.649 & 0.749 & 0.463 & 0.734 & 0.304 \\
BindWeave \cite{li2026bindweave}    & 0.471 & 0.970 & 0.610 & 0.635 & 0.399 & 0.557 & 0.111 \\ 
FFGO \cite{chen2025first}           & 0.413 & 0.939 & 0.673 & 0.580 & 0.422 & 0.739 & 0.235 \\
\midrule
UniVideo \cite{li2026bindweave}     & 0.463 & \underline{0.995} & 0.684 & 0.772 & 0.493 & 0.872 & 0.325 \\
VINO \cite{chen2026vino}            & 0.506 & 0.969 & \underline{0.694} & 0.768 & 0.509 & 0.868 & 0.331 \\
\midrule
% Omni-VCus \cite{li2025bindweave}    & - & - & - & - & - & - & - \\ \midrule
\rowcolor{yellow!30} Ours (Wan2.1-14B) & 0.508 & 0.994 & 0.691 & \underline{0.786} & \textbf{0.543} & \textbf{0.891} & \textbf{0.452} \\ 
\rowcolor{yellow!30} Ours (Wan2.2-14B) & \underline{0.525} & \textbf{0.997} & \textbf{0.696} & 0.760 & 0.537 & \underline{0.877} & \underline{0.431} \\ \bottomrule
\end{tabular}}
% \vspace{-4pt}
\label{tab:s2v_result}
\end{table*}
\section{Experiments}
\label{sec:experiment}
% ============ MAIN FIGURE HERE!!! ===============
% HoCo outperforms existing methods 
\begin{figure*}[t]
    \includegraphics[width=1\linewidth]{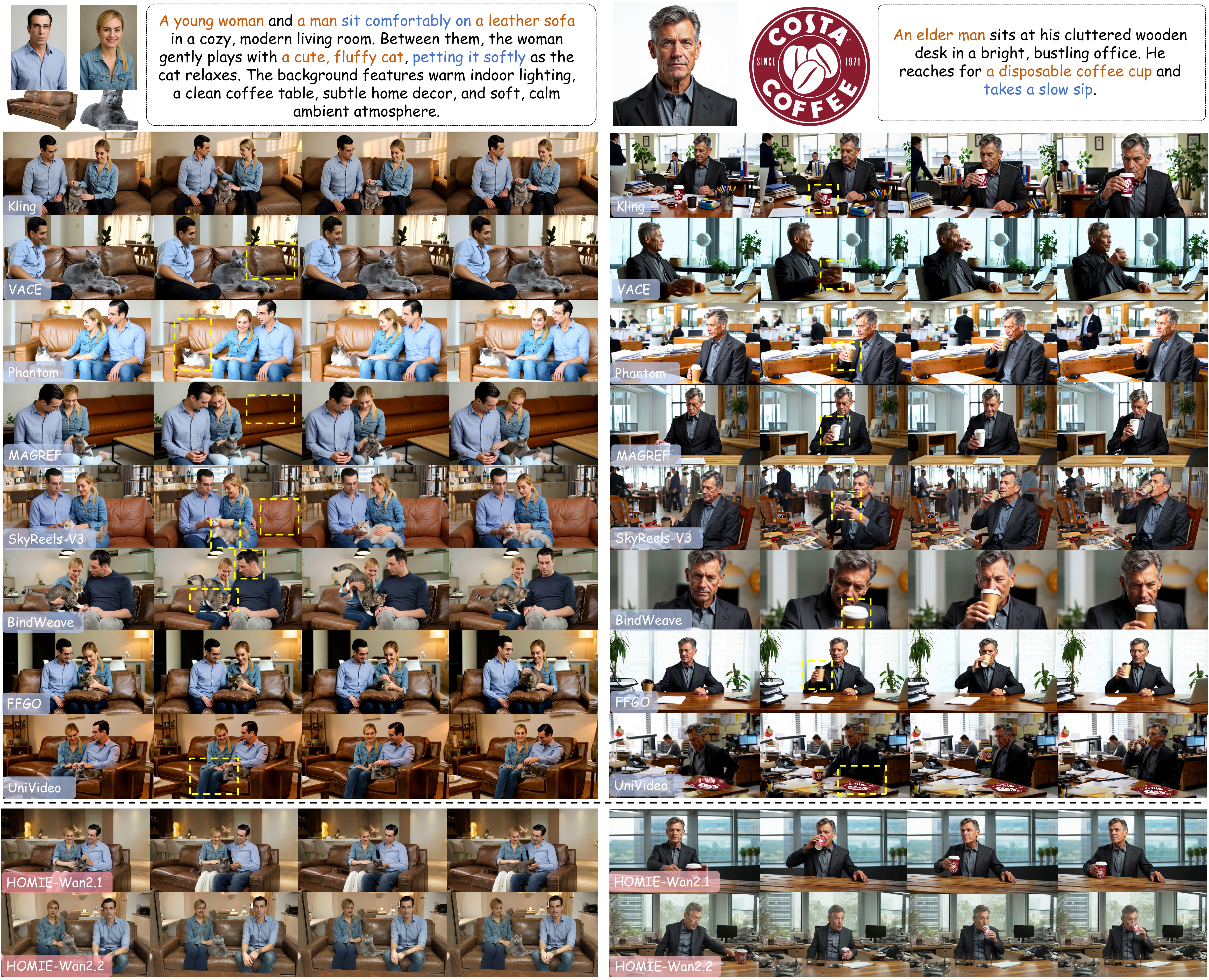}
    \caption{\textbf{Inter-subject} qualitative comparison between HOMIE and previous SOTA methods, including more subjects and abstract content personalization. Prompts that are relevant to subjects and interaction are highlighted. (\textit{Zoom in for the best view})}
    \label{fig:main_results}
\vspace{-5pt}
\end{figure*}

\begin{figure*}[htb]
    \includegraphics[width=1\linewidth]{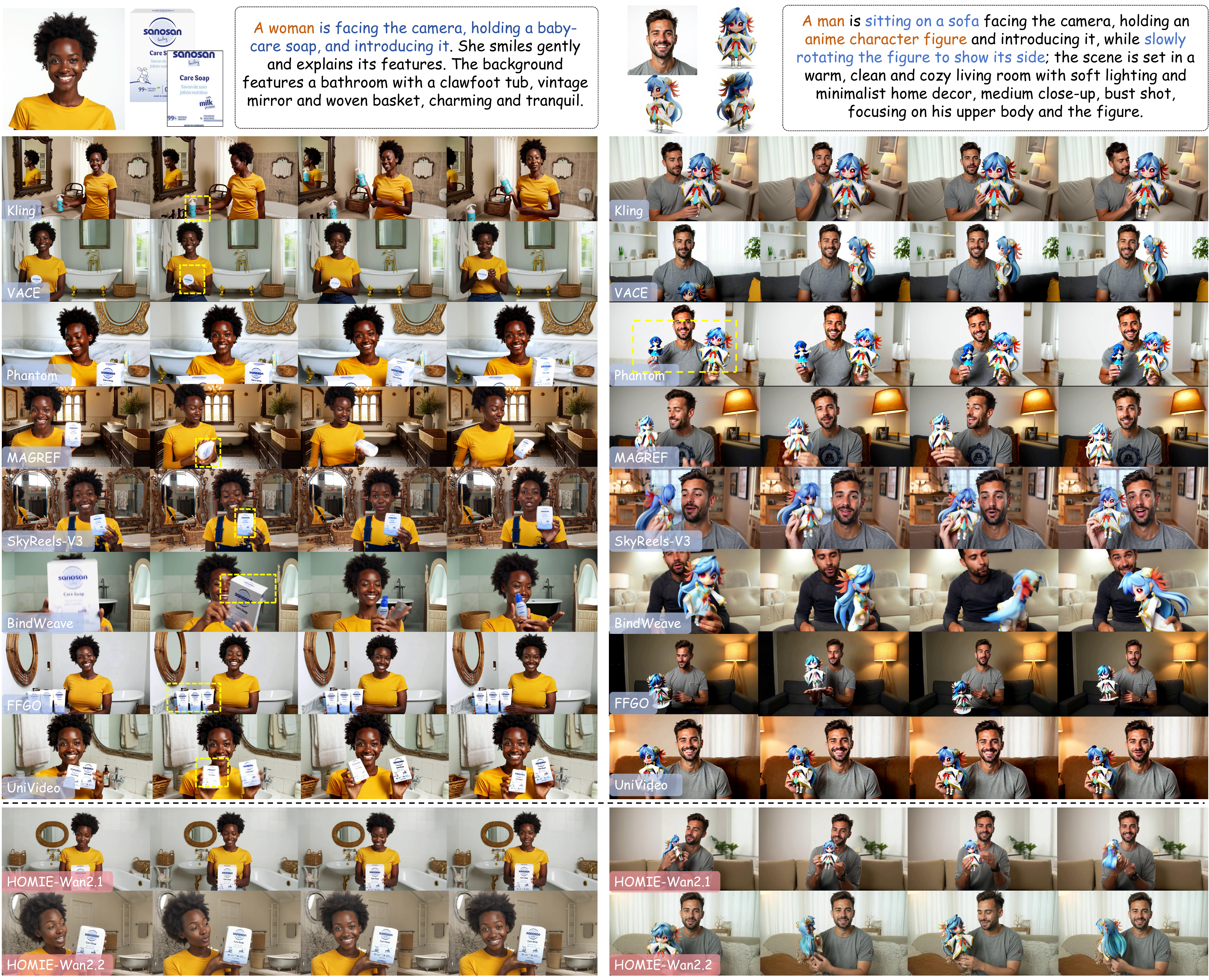}
    \caption{\textbf{Intra-subject} qualitative comparison between HOMIE and previous SOTA methods. Two representative intra-subject exemplars (OCR maps and multi-view references) are provided. Prompts that are relevant to subjects and interaction are highlighted. (\textit{Zoom in for the best view})}
    \label{fig:main_results_supp}
\vspace{-10pt}
\end{figure*}
% ============ MAIN FIGURE HERE!!! ===============

\noindent{\textbf{Implementation Details.}} We train HOMIE on the open-source text-to-video (T2V) foundation models Wan2.1-14B and Wan2.2-14B. We use Qwen3-VL-2B-Thinking to extract MLLM features from the last hidden state and employ a two-layer MLP to align them with 3D VAE tokens. GMG is inserted into all DiT blocks, and the number of reference embeddings is set to 5. During training, we adopt a three-stage tuning paradigm. In total, training consumes approximately 11K GPU-hours on A100 GPUs. Details of model configurations and training strategy are provided in Appendix.~\ref{sec:supp_implementation}.

\noindent{\textbf{Baselines.}}
We evaluate HOMIE against diverse baselines, including the closed-source Kling-1.6 \cite{kling} and multiple Wan-14B-based open-source methods: VACE-14B \cite{vace}, MAGREF \cite{deng2026magref}, SkyReels-A2 \cite{fei2025skyreels}, SkyReels-V3\cite{li2026skyreels}, Phantom \cite{phantom}, HuMo \cite{chen2026human}, BindWeave \cite{li2026bindweave}, and FFGO \cite{chen2025first}. To further validate its advantage compared with architectures replacing UmT5 with MLLMs (Fig.\ref{fig:mllm_design}(b)), we additionally evaluate against UniVideo\cite{wei2026univideo} and VINO\cite{chen2026vino}, which utilize the HunyuanVideo-T2V-13B backbone.

\noindent{\textbf{Evaluation.}}
We evaluate our method on a self-curated dataset of 200 human-object combinations, including 140 samples with more than two reference images, including diverse human identities and object categories in both inter-subject and intra-subject settings. Object images are sampled from open-source datasets \cite{ruiz2023dreambooth, yuan2026opensvnexus, xiang2025structured, huang2025adahuman}, while human reference images are drawn from both open-source data and AIGC-generated sources \cite{bflflux}. To ensure a comprehensive assessment, we employ three metric categories: (1) general video quality, quantified by aesthetic and motion smoothness scores \cite{yuan2026opensvnexus}; (2) text-following capability, measured via GMEScore \cite{gme}; and (3) subject consistency, evaluated using face similarity, DINO-I \cite{dinov2}, alongside GPT-5.2-based \cite{gpt52} object similarity and OCR accuracy. Additionally, for intra-subject with multi-view reference, we use $\mathrm{DINO_{rec}}$ and $\mathrm{DINO_{acc}}$ to evaluate if generated videos faithfully generate all viewpoints included in the references. Detailed experimental configurations are provided in Appendix.\ref{sec:supp_eval}.
\begin{wraptable}{r}{6.2cm}
\caption{Quantitative results of intra-subject HOCVP with multi-view references.}
\small
\label{tab:multiview_results}
\begin{tabular}{ccc}\toprule  
Method & $\mathrm{DINO}_{rec}$ & $\mathrm{DINO}_{acc}$ \\\midrule
VACE                               & 0.642 & 0.551 \\ 
Phantom                            & 0.633 & 0.553 \\ 
BindWeave                          & 0.614 & 0.546 \\
SkyReels-V3                        & 0.654 & 0.558 \\ \midrule
\rowcolor{yellow!30} Ours (Wan2.1) & 0.685 & \textbf{0.582} \\ 
\rowcolor{yellow!30} Ours (Wan2.2) & \textbf{0.696} & 0.560 \\\bottomrule
\end{tabular}
\vspace{-35pt}
\end{wraptable}

\subsection{Main Results}
\noindent{\textbf{Quantitative Comparison.}}

As detailed in Tab.\ref{tab:s2v_result}, HOMIE achieves state-of-the-art video quality and text-following performance among open-source methods, while maintaining competitive subject consistency across multiple metrics under diverse reference settings. Notably, HOMIE shows 21.8\% improvement in OCR accuracy (relative to SkyReels-V3), validating its effective use of OCR maps from the evaluation dataset. Moreover, as reported in Tab. \ref{tab:multiview_results}, HOMIE also delivers competitive performance in multi-view intra-subject HOCVP, demonstrating its strong ability to faithfully synthesize diverse views of the same object while maintaining their corresponding subject fidelity.

\noindent{\textbf{Qualitative Comparison.}} Fig.\ref{fig:main_results} presents qualitative comparisons between HOMIE and various baselines on challenging inter-subject HOCVP tasks. Overall, HOMIE achieves an optimal balance between subject consistency and text following. In a complex 4-subject HOCVP example, VACE, Skyreels-V3, BindWeave, and UniVideo struggle with subject consistency (exhibiting missing subjects, incorrect attire, or body degradation), while Kling exhibits less fluent motion. For logo personalization, which requires reasoning capabilities, only HOMIE, Kling, and UniVideo successfully attach the \textit{COSTA} logo to the \textit{disposable paper cup}, \textbf{without explicitly mentioning such a relationship in the prompt}. However, only HOMIE faithfully renders the logo details. Fig.\ref{fig:main_results_supp} displays comparisons on intra-subject HOCVP tasks. On the one hand, HOMIE best maintains textual information, while Phantom, FFGO, and UniVideo produce replicated generations, indicating the models mistakenly treat the OCR map as a distinct subject. On the other hand, in multi-view HOCVP instances, HOMIE shows a noticeable advantage in maintaining diverse perspectives of the object while following the interaction pattern of \textit{rotating}. Additional comparison results and qualitative examples are provided in Appendix.~\ref{sec:supp_extend_comparison},~\ref{sec:supp_more_result}. All video samples can be found on our demo webpage.

\subsection{Ablation Study}
We present an ablation study on our model design, with results detailed in Tab.~\ref{tab:ablation}. The ``Plain'' denotes a configuration that omits all multimodal inputs (along with the GMG) and the MRE modules. We also conduct experiments by ablating the GMG and MRE components individually. Notably, the ``MLLM to UmT5'' denotes the integration of MLLM features into the UmT5 embeddings, which corresponds to the architecture illustrated in Fig.~\ref{fig:mllm_design}(a) utilizing an identical alignment network.

\begin{table}[t]
\centering
\small
\caption{Ablation study of various designs and components.}
\label{tab:ablation}
\resizebox{0.98\columnwidth}{!}{%
\begin{tabular}{lccccc}
\hline
\multicolumn{1}{c}{Setting} & GMEScore $\uparrow$ & Face-Sim $\uparrow$ & DINO-I $\uparrow$ & Obj-Sim $\uparrow$ & OCR Acc. $\uparrow$ \\ \hline
a) Plain (No MLLM)     & 0.682 & 0.741 & 0.506 & 0.853 & 0.388 \\
b) w/o GMG             & 0.688 & 0.697 & 0.502 & 0.836 & 0.430 \\
c) w/o MRE             & 0.690 & 0.738 & 0.508 & 0.840 & 0.376 \\
d) MLLM to UmT5        & 0.675 & 0.725 & 0.516 & 0.863 & 0.422 \\\midrule
e) Ours (Wan2.1-14B)   & \textbf{0.691} & \textbf{0.786} & \textbf{0.543} & \textbf{0.891} & \textbf{0.452} \\\hline
\end{tabular}%
}
\end{table}

\begin{figure*}[t]
    \includegraphics[width=0.95\linewidth]{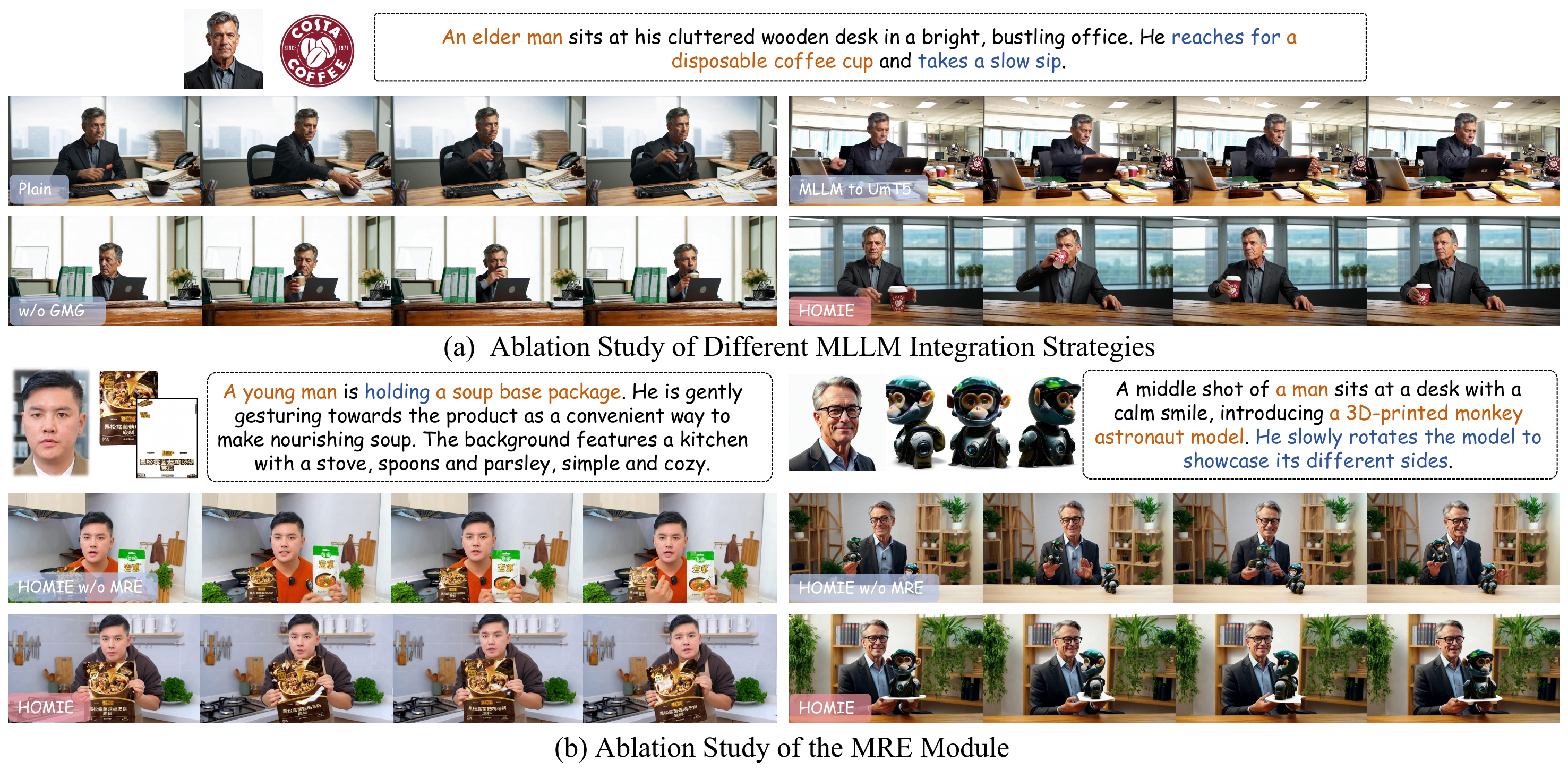}
    \caption{Qualitative Ablation Studies (\textit{Zoom in for the best view}.)}
    \label{fig:ablation}
\vspace{-5pt}
\end{figure*}

\noindent{\textbf{Different Integration Strategies of MLLM Feature.}}
Fig.~\ref{fig:ablation}(a) qualitatively compares strategies for integrating MLLM features into video models. For the challenging task of logo personalization, only GMG successfully links logos to semantically relevant objects, demonstrating its effective use of MLLM knowledge for reasoning. In contrast, integrating these features into UmT5 fails to establish such links or generate the \textit{taking a sip} motion, consistent with the GMEScore drop in Setting (d) of Tab.~\ref{tab:ablation}. This matches our analysis regarding the limitations of aligning MLLM features with UmT5. 

\begin{wrapfigure}{r}{0.48\textwidth}
\vspace{-15pt}
\includegraphics[width=0.47\textwidth]{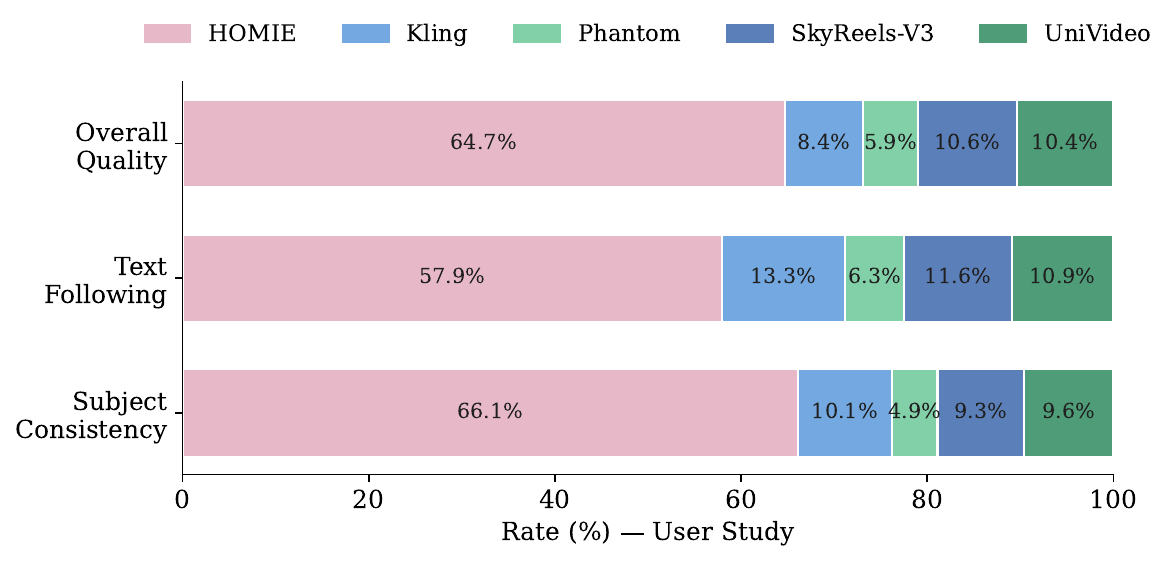}
\caption{User study results.}
\vspace{-25pt}
\label{fig:user_study}
\end{wrapfigure}

\noindent{\textbf{Modality-Reference Embedding.}}
Fig.~\ref{fig:ablation}(b) presents ablation results for the MRE module across two intra-subject HOCVP scenarios. Without MRE, the OCR map and multi-view inputs from the original reference are treated as independent entities, causing their unintended appearance in the generated video. This confirms that MRE effectively captures the intrinsic semantic links among the reference images.

\subsection{User Study}
In a user study with 40 participants, we assess the overall quality, text following, and subject consistency of 20 samples, each containing five videos generated by HOMIE, Phantom, Kling, SkyReels-V3, and UniVideo. Participants were asked to select the best video for each metric. As shown in Fig.~\ref{fig:user_study}, HOMIE consistently outperforms all baselines, highlighting its advantages for real-world deployment. Detailed user study guidelines are provided in Appendix.~\ref{sec:supp_eval}.

%% file: 5_conclusion.tex
\section{Conclusion}
\label{sec:conclusion}
In this paper, we propose HOMIE, a framework that successfully integrates knowledge from MLLMs to improve HOCVP performance. Unlike prior approaches, we introduce a multimodal input paradigm paired with global multimodal guidance, thereby enhancing MLLM feature interactions within the video diffusion model. To further tackle scenarios that include intra-subject references, we design learnable modality-reference embeddings that help the model distinguish token modalities and capture internal connections among references. Comprehensive experiments demonstrate that HOMIE achieves state-of-the-art results across diverse HOCVP tasks, especially in challenging scenarios such as logo personalization and multi-view video generation of specific objects.

%% file: x_supp.tex
\section{Appendix}

Our supplementary material is organized into the following sections:
\begin{itemize}[leftmargin=2.0em]
\item Section~\ref{sec:supp_implementation} provides details of our experimental setup, including HOCVP training strategies. For reference, we report the training costs of HOMIE and its counterparts.
\item Section~\ref{sec:supp_eval} reports the details of our evaluation, including the comparable baselines, evaluation datasets, and metrics. Additionally, we provide the guidelines for the user study.
\item Section~\ref{sec:supp_extend_comparison} presents additional qualitative comparisons and corresponding discussions across diverse tasks, such as complex inter-subject and intra-subject personalization. We specifically include comparisons with MLLM-integrated methods, namely SkyReels-A2, BindWeave, UniVideo, and VINO.
\item Section~\ref{sec:supp_more_result} showcases further generation results from HOMIE, together with cases for practical applications.
\item Section~\ref{sec:more_discussion} offers further discussion on abstract concept personalization, ethical considerations, and the potential limitations of our work.
\end{itemize}

The original video samples for all examples mentioned in both the main submission and the supplementary materials are provided on our demo webpage: \url{https://homie-demo.github.io/}.

\subsection{Implementation Details}
\label{sec:supp_implementation}
\subsubsection{Model Designs} 
\noindent \textbf{MLLM feature generation.} We extract the last hidden state of Qwen3-VL-2B-Thinking in advance during training. The system prompt $\mathcal{P}_{sys}$ mentioned in Sec.~\ref{sec:input_paradigm} during training and inference is kept identical as Tab.\ref{tab:sys_prompt}:

\begin{table*}[h]
\centering
\caption{System prompt of MLLM feature extraction.}
\label{tab:sys_prompt}
\small
\resizebox{1.0\textwidth}{!}{%
\begin{tabular}{p{0.95\textwidth}}
\toprule
1. You are an expert multimodal AI assistant specialized in highly controllable visual content generation. \\
 
2. You will receive A set of images depicting specific humans and objects and a text description detailing the target patterns of human-object interaction. \\

3. Your task: analyze the precise relationships between the visual entities in the images and the interaction logic defined in the text prompt. Specifically, evaluate how the distinct human and object features from the visual inputs should be spatially, physically, and semantically integrated to accurately reconstruct the interaction described in the text. \\
 \bottomrule
\end{tabular}%
}
\end{table*}

% ===== 

The output hidden states are kept at a maximum length of 1024 and a hidden state dimension of 2048.

\noindent \textbf{Global Multimodal Guidance.} To ensure sufficient interaction between multimodal features and video tokens, we apply GMG to all self-attention layers across the 40 DiT blocks in the Wan model.

\subsubsection{Training Strategies}
% refined 
As noted in Sec.\ref{sec:dataset_curation} of the main paper, we construct a dataset consisting of three components: a single-reference video dataset, a multiple-reference video dataset, and a high-resolution multiple-reference video dataset. To enable the model to incrementally acquire capabilities across varying levels of task difficulty, we propose a multi-stage training strategy, the training configurations of which are reported in Tab. \ref{tab:training_details}.

\begin{table}[h]
\centering
\caption{Training details across different stages}
\label{tab:training_details}
\begin{tabular}{lccc}
\toprule
\textbf{Training Details} & \textbf{Stage 1} & \textbf{Stage 2} & \textbf{Stage 3} \\
\midrule
Learning Rate & 1e-5 & 1e-5 & 2e-6 \\
Beta $(\beta_1, \beta_2)$ & (0.9, 0.999) & (0.9, 0.999) & (0.9, 0.999) \\
Training Steps & 3000 & 2000 & 500 \\
Resolution & $480 \times 832$ & $480 \times 832$ & $720 \times 1280$ \\
Reference & Single & Multiple & Multiple \\
\bottomrule
\end{tabular}
\end{table}

\noindent{\textbf{Single-Subject Training}}
In this phase, we train our model on the single-subject-driven video dataset with 300K samples. The core objective here is to learn diverse subject representations while preserving high identity consistency. To this end, we randomly select one sample from the reference embeddings and assign it to the single reference. This strategy ensures that each reference embedding is sufficiently optimized and uniquely associated with distinct subject identities.

\noindent{\textbf{HOC Training}}
After the first training stage, the model already achieves robust identity consistency across diverse subject categories. To further enhance its ability to capture semantic knowledge, such as interaction patterns between specific subjects and the object most relevant to a given logo, we continue training on a multiple-reference video dataset. During this stage, the selection of reference embeddings takes a fixed order, and the inference process follows this order as well.

\noindent{\textbf{High-Resolution Training}}
The first two stages use uniformly formatted 480P video data. For the final stage, we adopt the filtered 720P multiple-reference dataset and conduct a short-term fine-tuning process to enhance the capacity for high-resolution video generation.

We also compare training costs with a subset of baseline models. Notably, methods such as UniVideo and VINO, which fully replace the original text encoder with MLLMs, require substantially more computational resources. This is because their entire control pipeline must undergo extensive re-alignment. In contrast, our design integrates an MLLM while retaining the original text encoder, which proves to be a more efficient strategy. By preserving the encoder’s basic textual control capabilities, the MLLM can focus on learning more advanced control mechanisms from diverse multimodal inputs.
\begin{table*}[h]
    \centering
    \caption{Comparison of training costs}
    \begin{tabular}{ccc}
    \toprule
      \textbf{Method}    & \textbf{Base Model} & \textbf{Training Cost (Summary)} \\ \midrule
      Phantom   & Wan2.1-14B &  30k A100 hours \\ \midrule
      VACE   & Wan2.1-14B & 200k steps on 128 A100 GPUs \\ \midrule
      BindWeave & Wan2.1-14B & 6k steps on 512 xPUs \\ \midrule
      UniVideo  & HunyuanVideo-13B & 3-stage training, totally 35k steps \\ \midrule
      VINO      & HunyuanVideo-13B & 3-stage training, totally 40k steps \\ \midrule
      HOMIE-Wan2.1 & Wan2.1-14B & 5.5k steps, 10k A100 hours \\
    \bottomrule
    \end{tabular}
    
    \label{tab:training_cost}
\end{table*}

\clearpage

\subsection{Evaluation Details}
\label{sec:supp_eval}

\begin{table*}[h]
    \centering
    \caption{Details of backbones and technical implementations compared to previous SOTAs.}
     \resizebox{1.0\textwidth}{!}{
    \begin{tabular}{cccccc c c}
        \toprule
       \textbf{Method}    & \textbf{Base Model Name} & \textbf{Base Model Type} &  \textbf{MLLM} & \textbf{Intra-subject Distinction}  \\
        \midrule
        Kling 1.6  & Closed source  & - & - &  - & - \\
        \midrule
        
        SkyReels-A2  & Wan2.1 (14B)  & T2V & \cmark & \xmark  \\    
        HuMo (AAAI'26)  & Wan2.1 (14B)  & I2V & \xmark & \xmark \\
        VACE-14B (ICCV'25)   & Wan2.1 (14B)  & T2V & \xmark & \xmark  \\
        Phantom (ICCV'25) & Wan2.1 (14B)   & T2V & \xmark & \xmark  \\
        BindWeave (ICLR'26) & Wan2.1 (14B)  & T2V & \cmark & \xmark  \\
        MAGREF (ICLR'26) & Wan2.1 (14B)  & I2V & \xmark & \xmark  \\
        FFGO (CVPR'26)  & Wan2.2 (14B)  & I2V & \xmark & \xmark  \\
        UniVideo (ICLR'26)  & HunyuanVideo (13B)  & T2V & \cmark & \xmark \\
        VINO (Arxiv'26)  & HunyuanVideo (13B)  & T2V & \cmark & \xmark  \\
        SkyReels-V3 (Arxiv'26)  & Wan2.1 (14B)  & T2V & \xmark & \xmark  \\
        \midrule
        Ours & Wan2.1\& Wan2.2 (14B)  & T2V & \cmark & \cmark \\
        \bottomrule
    \end{tabular}}
    \label{tab:comparison}
\end{table*}
\subsubsection{Discussion on Prior SOTA Methods}
As shown in Tab.\ref{tab:comparison}, we present the details of previous state-of-the-art methods, which are included in our comparison experiments. We discuss:

\noindent{\textbf{Base models.}} Except for Kling-1.6, which is a closed-source commercial model, most methods are built on Wan-14B models. To ensure a relatively fair comparison, we follow this setting and train two versions of HOMIE based on Wan2.1-T2V-14B and Wan2.2-T2V-14B. To further highlight the advantages of our method, we also compare against UniVideo and VINO, which are built on HunyuanVideo and replace the original text encoders with MLLMs. However, their training pipelines—such as substituting the original encoders with QwenVL-2.5-7B—incur substantially higher costs, while also achieving stronger performance than their Wan-14B-based counterparts, including HOMIE.

\noindent{\textbf{Technical Implementation.}} ``MLLM'' indicates whether a multimodal large language model is incorporated into the training of video personalization to enhance generative performance. ``Intra-subject Distinction'' indicates whether the method's implementation can identify inter- and intra-subject references. HOMIE's GMG and MRE modules fit both aspects and show competitive performance across a broad range of scenarios.

\subsubsection{Evaluation Datasets}
We self-curate an evaluation dataset containing 200 human-object-centric samples. We follow the steps below to construct our evaluation set: 

\noindent{\textbf{Step I: Collection of Reference Images.}} We collect human images from the OpenS2V-Eval benchmark \cite{yuan2026opensvnexus}. To increase human diversity, we also utilize FLUX.1-dev \cite{bflflux} to synthesize a set of human portraits with diverse attributes, including variations in gender, hairstyle, race, and age. For object references, we collect samples from OpenS2V-Eval and DreamBooth \cite{ruiz2023dreambooth}. Furthermore, we incorporate product images containing abundant text information (paired with OCR maps extracted by state-of-the-art OCR models), logos (for evaluating logo personalization), and multi-view object images sampled from Trellis \cite{xiang2025structured} and AdaHuman \cite{huang2025adahuman}. 

\noindent{\textbf{Step II: Construction of the HOCVP Evaluation Dataset.}} Every sample contains at least one human image and one object image. We use MLLMs to generate the corresponding prompts following OpenS2V-Eval. Notably, we ensure that 140 out of the 200 samples contain three or more reference images. In terms of scenarios, 100 samples are purely inter-subject HOCVP samples, including 25 samples for abstract concept personalization. For the remaining 100 samples, 60 contain intra-subject references to OCR maps, and 40 contain multi-view images. Through this pipeline, we establish a robust evaluation dataset that covers comprehensive tasks alongside diverse human and object subjects.

We will release all samples in the evaluation dataset later.

\subsubsection{Evaluation Metrics}
\noindent{\textbf{Video Quality.}} We select aesthetic score and motion smoothness from the OpenS2V-Nexus benchmark \footnote{\url{https://github.com/PKU-YuanGroup/OpenS2V-Nexus/tree/main/eval}} to evaluate the video quality generated by each baseline.

\noindent{\textbf{Text following.}} 
We use GMEScore to evaluate the text following of generated videos and prompts. 

\noindent{\textbf{Subject Consistency.}}
We evaluate the subject consistency of each method using four metrics: face similarity, DINO-I similarity, object similarity, and OCR accuracy. 

\begin{itemize}
    \item \textbf{Face Similarity.} Following the common implementation of existing methods, we use a face feature extractor (\textit{i.e.}, CurricularFace) to extract facial features from the generated video. After sampling the same number of frames for each method, we calculate the average cosine similarity between the generated faces in the video and the face in the reference image.
    \item \textbf{DINO-Score.} We employ Grounded-SAM-2\footnote{\url{https://github.com/IDEA-Research/Grounded-SAM-2}} to extract target objects from sampled frames. We then calculate the cosine similarity between DINO-v2 embeddings of reference objects and the segmented objects.
    \item \textbf{Object Similarity.} We use GPT-5.2 to assign a similarity score between sampled frames and reference images.
    \item \textbf{DINO-Score for Multi-view Reference.} In HOCVP tasks involving intra-subject multi-view references, ensuring bidirectional fidelity is essential: each reference viewpoint should be accurately generated, and every video frame should consistently depict the high-fidelity object. To evaluate this requirement, following \cite{songmv}, we utilize the DINO recall $\mathrm{DINO}_{rec}$ and DINO accuracy $\mathrm{DINO}_{acc}$. Given $R$ multi-view reference images $\{I_{r}^{ref}|r = 1\cdots R\}$ and $V$ sampled frames $\{I_{v}^{vid}|v = 1\cdots V\}$ from the generated video, we define :
    \begin{equation}
        \mathrm{DINO}_{rec} = \frac{1}{R} \sum_{r=1}^{R} \underset{v\in\{1,\cdots V\}}{max} \mathrm{DINO}(I^{ref}_r, I^{vid}_{v})
    \end{equation}
    \begin{equation}
        \mathrm{DINO}_{acc} = \frac{1}{V} \sum_{v=1}^{V} \underset{r\in\{1,\cdots R\}}{max} \mathrm{DINO}(I^{ref}_r, I^{vid}_{v})
    \end{equation}
    \item \textbf{OCR Accuracy.} We leverage GPT-5.2 to detect text content from reference images (treated as ground truth) and compute the edit distance\footnote{\url{https://pypi.org/project/python-Levenshtein/}} against OCR results extracted from sampled frames. Let $S_{\text{ref}}$ and $S_{\text{pred}}$ denote the OCR recognition results extracted from the reference image and a sampled frame, respectively. We adopt the normalized Levenshtein similarity to quantify the OCR accuracy, which is formulated as:
    \begin{equation}
        \text{sim}_{\text{norm}}(S_{pred}, S_{ref}) = 1 - \frac{\text{ed}(S_{pred}, S_{ref})}{\max\left\{ |S_{pred}|, |S_{ref}| \right\}}.
    \end{equation}
    Here, $|S|$ indicates the character sequence length of a string $S$, $\text{ed}(\cdot,\cdot)$ denotes the standard Levenshtein edit distance, and the normalized similarity $\text{sim}_{\text{norm}} \in [0,1]$, where a value of 1 corresponds to a perfect match between $S_{\text{pred}}$ and $S_{\text{ref}}$.
\end{itemize}

\begin{table*}[h]
\centering
\caption{Prompts input to GPT-5.2 to obtain object similarity scores and OCR results.}
\label{tab:prompt_gpt}
\resizebox{1.0\textwidth}{!}{
\begin{tabular}{ll}
        \toprule
        Metric & Prompt   \\
        \midrule
        Object Similarity Scores  & \makecell[l]{
            You are an AI visual similarity evaluator specializing in product consistency assessment. You will receive two images: \\ 
            Image 1: A standalone product (the reference item). \\ 
            Image 2: A person holding a product (the target item to be evaluated). \\ 
            Your task is to quantify the visual similarity between the product in Image 1 and the product in Image 2 \\based on core attributes including
            shape, color, texture, brand features, and overall appearance. \\ 
            Scoring Rules: \\
            1. Assign a score between 0 and 10 (inclusive, can be a decimal value, e.g., 4.5). \\
            2. Metrics: \\
            \tabitem $[8,10]$: Identical to reference in shape, texture, color, size, and fine-grained details like logos and text; \\
            \tabitem $[6,8)$: Key features fully preserved; only negligible, non-critical visual deviations; \\
            \tabitem $[4,6)$: Same category as reference; noticeable but minor mismatches in secondary details; \\
            \tabitem $[2,4)$: Category-aligned; severe distortions in key features, leading to semantic ambiguity; \\
            \tabitem $[0,2)$: Minimal reference cues; object category barely inferable from generated content; \\
            3. Higher scores correspond to higher similarity. \\
            Strict Output Requirement: Return only the numerical score (no explanations, text descriptions, or additional comments).
            % \begin{itemize}
            %     \item $[8, 10]$: Identical to reference in shape, texture, color, size, and fine-grained details like logos and text;
            %     \item $[6, 8)$: 
            % \end{itemize}
        }   \\ \midrule
        OCR Results  & \makecell[l]{You are an AI specialized in image OCR text extraction. \\
Analyze the provided image and locate the product held by the person in the image, extract all text content on this product. \\
Output only the raw OCR text results without any additional explanations, notes, or formatting.} \\
        \bottomrule
\end{tabular}}
\end{table*}

\begin{table*}[t]
\centering
\caption{Guidelines of user study.}
\label{tab:user_study}
\small
\resizebox{1.0\textwidth}{!}{%
\begin{tabular}{p{0.95\textwidth}}
\toprule
Guidelines: Subject-driven video personalization is an important downstream task of text-based video generation, aiming to generate videos based on user-provided images and prompts.  Please watch the following videos of a person holding an object, generated from a reference person image, a reference object image, and a prompt. Compare their effects and evaluate the generated video based on three metrics:\\
\midrule
1. \textbf{Video Quality}: Evaluate video quality based on the smoothness of the person's movements (avoiding static video of the person holding an object), the realism of texture (normal color and saturation), and physical consistency (objects appearing suspended, clipping). \\
 
2. \textbf{Text Following}: Evaluate text following based on the consistency between the generated video and the input text description (e.g., the person's behavior). \\

3. \textbf{Subject Consistency}: Evaluate subject consistency based on the similarity between the generated person or object and the reference person or object image, respectively (e.g., appearance, color, shape), and the reasonable placement of the reference object (e.g., an abstract concept logo). \\
\midrule

Please rank the baselines and our method across these three metrics. \\
 \bottomrule
\end{tabular}%
}
\end{table*}

The overall metric score of a generated video is computed as the average of scores extracted from all sampled frames. To ensure fair comparison, we uniformly sample 16 equidistant frames from each video. The prompts used to guide GPT-5.2 in returning object similarity scores and OCR results are provided in Tab. \ref{tab:prompt_gpt}.

\subsubsection{Evaluation Guidance of User Study}
As shown in Tab. \ref{tab:user_study}, we present the evaluation guidance for the user study. Users rate the counterparts and our method based on three metrics: Video Quality, Text Following, and Subject Consistency.

\subsection{Extended Qualitative Comparisons}
\label{sec:supp_extend_comparison}
\begin{figure*}[htbp]
    \includegraphics[width=1.0\linewidth]{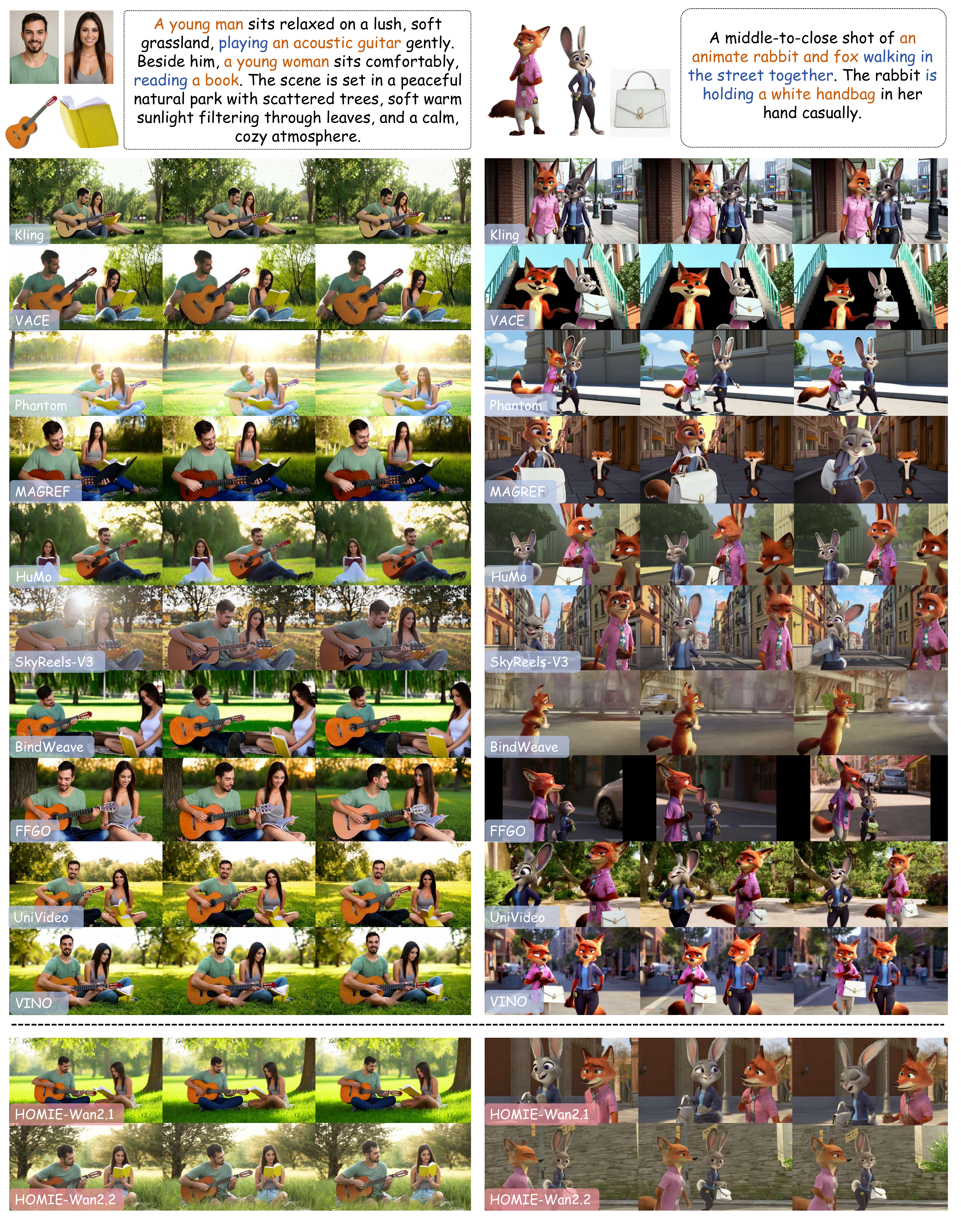}
    \caption{More \textbf{inter-subject} qualitative comparison between HOMIE and previous SOTA methods, including more subjects and complex interaction patterns. Prompts that are relevant to subjects and interaction are highlighted. (\textit{Zoom in for the best view})}
    \label{fig:supp_multi_1}
\end{figure*}

\begin{figure*}[htbp]
    \includegraphics[width=1.0\linewidth]{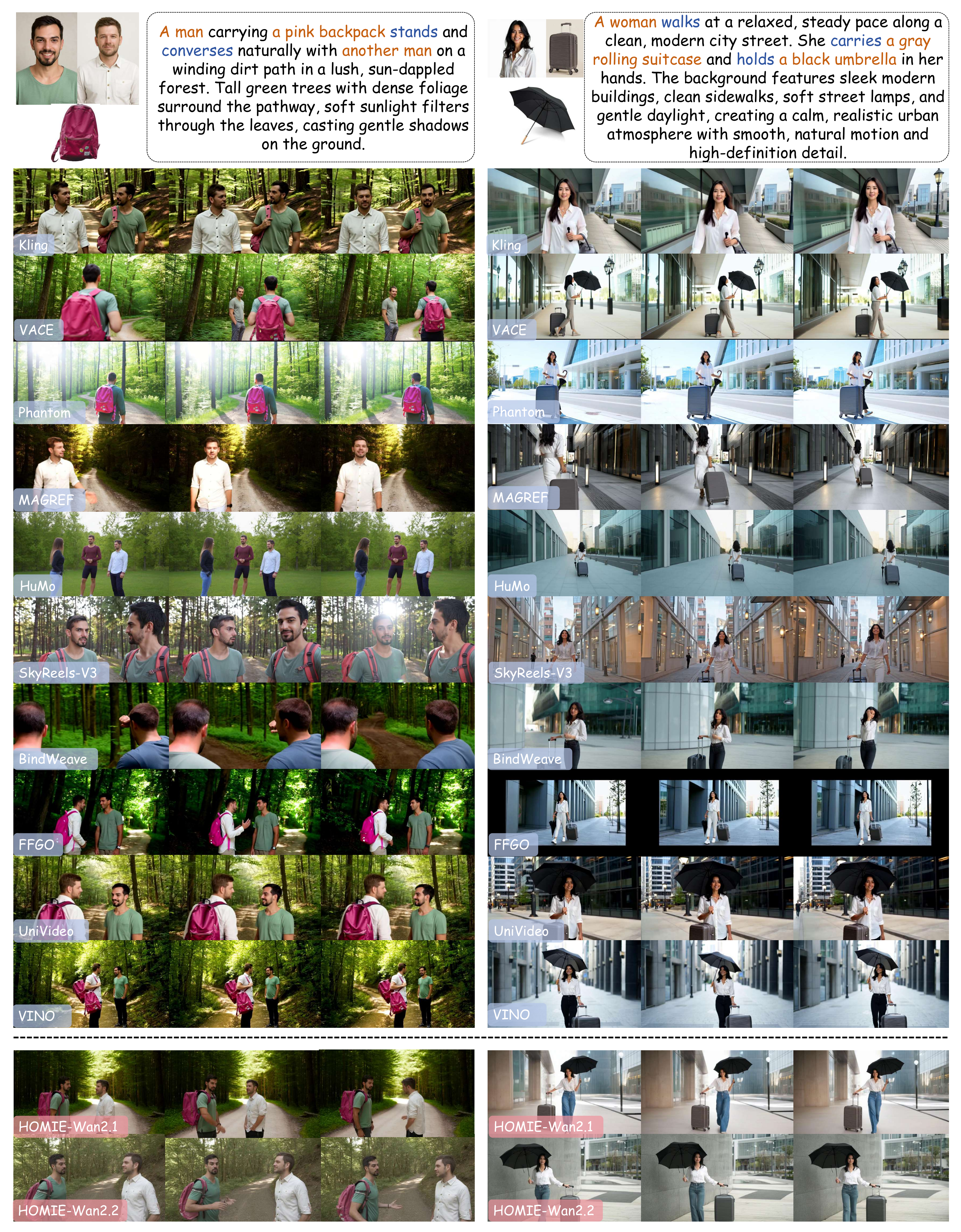}
    \caption{More \textbf{inter-subject} qualitative comparison between HOMIE and previous SOTA methods, including more subjects and complex interaction patterns. Prompts that are relevant to subjects and interaction are highlighted. (\textit{Zoom in for the best view})}
    \label{fig:supp_multi_2}
\end{figure*}

\begin{figure*}[htbp]
    \includegraphics[width=1.00\linewidth]{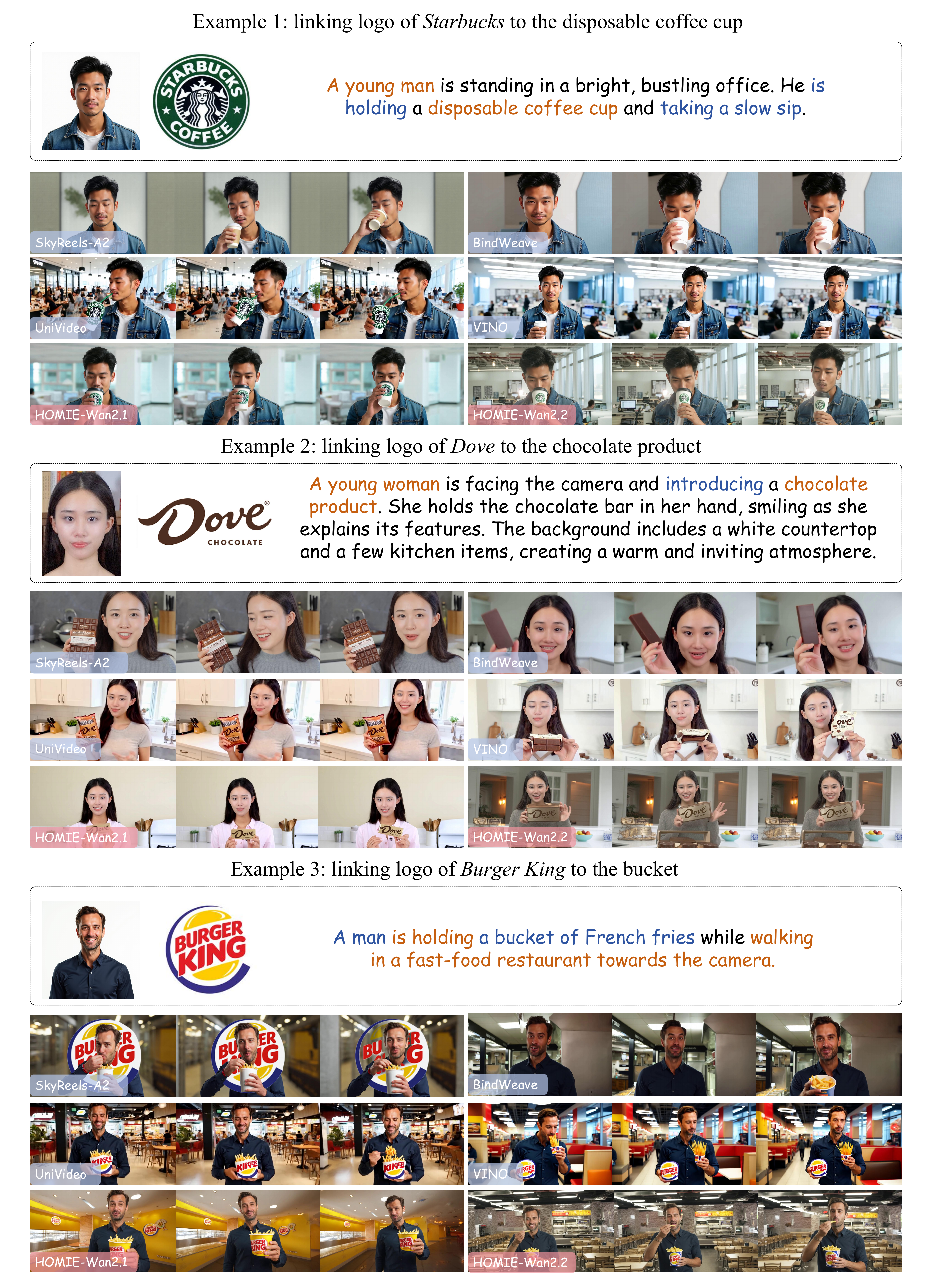}
    \caption{More qualitative comparison between HOMIE and previous SOTA methods (integrated MLLMs) on abstract concept (\textit{e.g.}, logo) personalization, a challenging inter-subject HOCVP scenario, which requires reasoning ability from MLLMs. (\textit{Zoom in for the best view})}
    \label{fig:supp_logo}
\end{figure*}

\begin{figure*}[htbp]
    \includegraphics[width=1.00\linewidth]{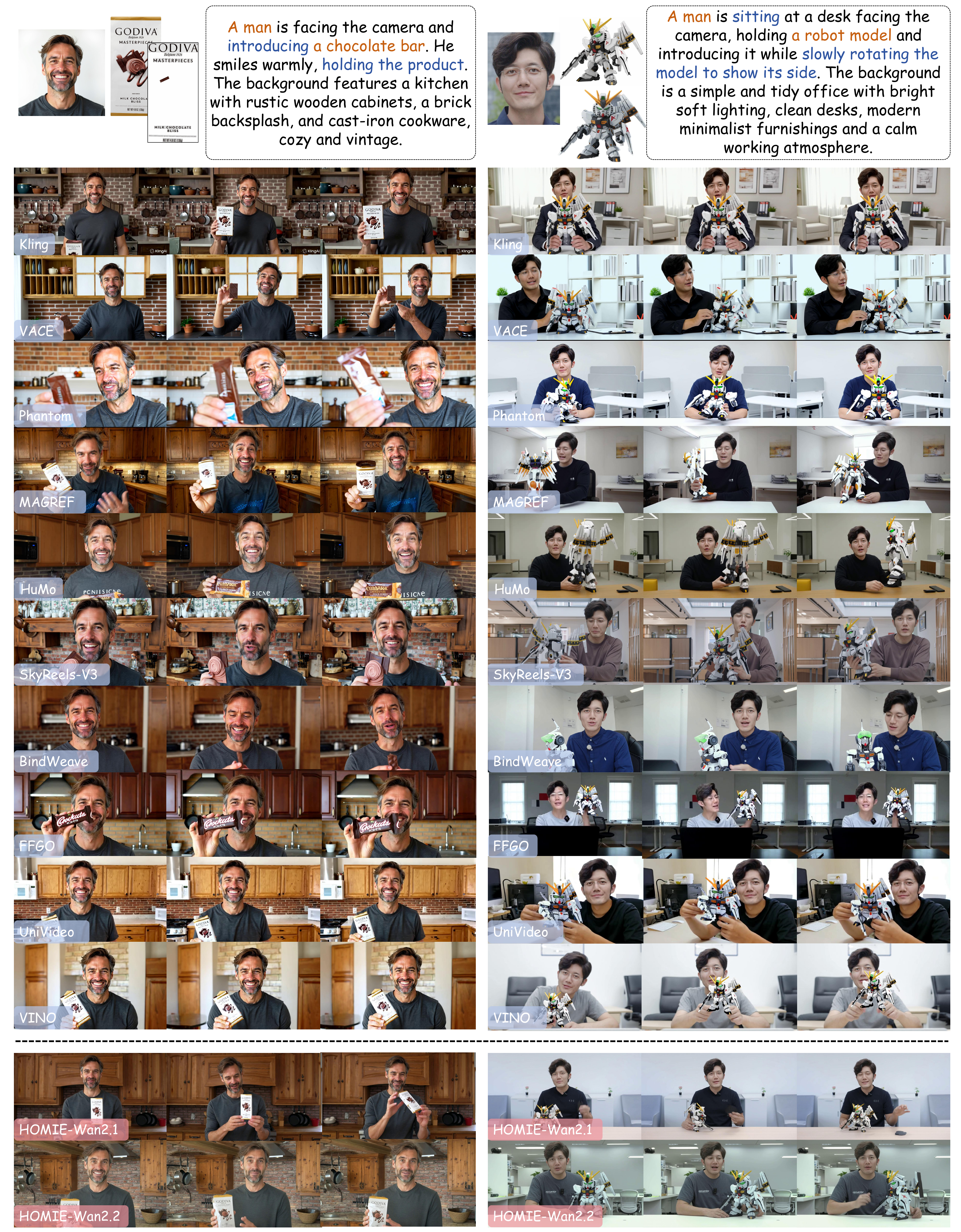}
    \caption{More \textbf{intra-subject} qualitative comparison between HOMIE and previous SOTA methods. Two representative intra-subject exemplars (OCR maps and multi-view references) are provided. Prompts that are relevant to subjects and interaction are highlighted. (\textit{Zoom in for the best view})}
    \label{fig:supp_intra}
\end{figure*}

In this section, we present additional comparisons with prior methods. Fig.~\ref{fig:supp_multi_1} and Fig.~\ref{fig:supp_multi_2} demonstrate challenging inter-subject HOCVP examples, including animate characters and diverse objects. These results show that HOMIE achieves competitive performance, whereas other methods struggle with incorrect subject counts, limited consistency, and unnatural interactions. Furthermore, Fig.~\ref{fig:supp_logo} illustrates challenging abstract concept personalization results compared to MLLM-integrated methods such as SkyReels-A2, BindWeave, UniVideo, and VINO. These comparisons solidify HOMIE's advantages in two key aspects: (1) achieving highly natural human-object interactions, which demonstrates its successful integration of MLLM knowledge, and (2) accurately rendering logos onto the most contextually relevant objects. Finally, Fig.~\ref{fig:supp_intra} demonstrates HOMIE's advantages in intra-subject HOCVP using subject images paired with OCR maps and multi-view inputs. By effectively leveraging these references, HOMIE significantly enhances both textual fidelity and multi-view consistency in the generated videos. \textbf{All comparison videos are available on our demo webpage.} 
% ============ MAIN FIGURE HERE!!! ===============

% ============ MAIN FIGURE HERE!!! ===============

% ============ MAIN FIGURE HERE!!! ===============

\clearpage

\subsection{More Results}
\label{sec:supp_more_result}
\begin{figure*}[h]
    \includegraphics[width=0.95\linewidth]{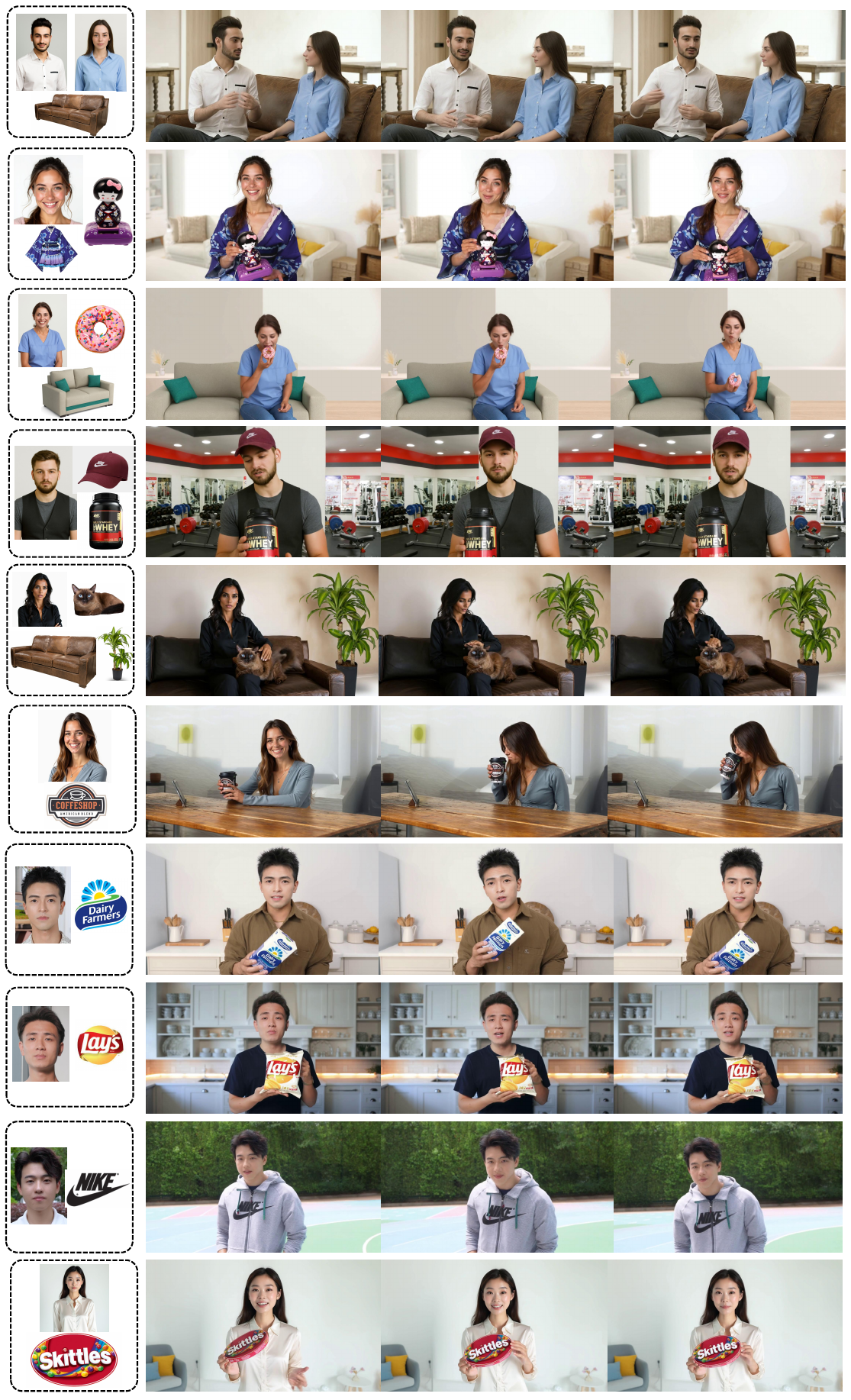}
    \caption{Additional visual results of HOMIE on inter-subject HOCVP scenarios, including general human-object-interaction personalization and abstract concept (\textit{e.g.}, logos) personalization.}
    \label{fig:supp_more_results_1}
\end{figure*}

\begin{figure*}[h]
\begin{center}
    \includegraphics[width=0.95\linewidth]{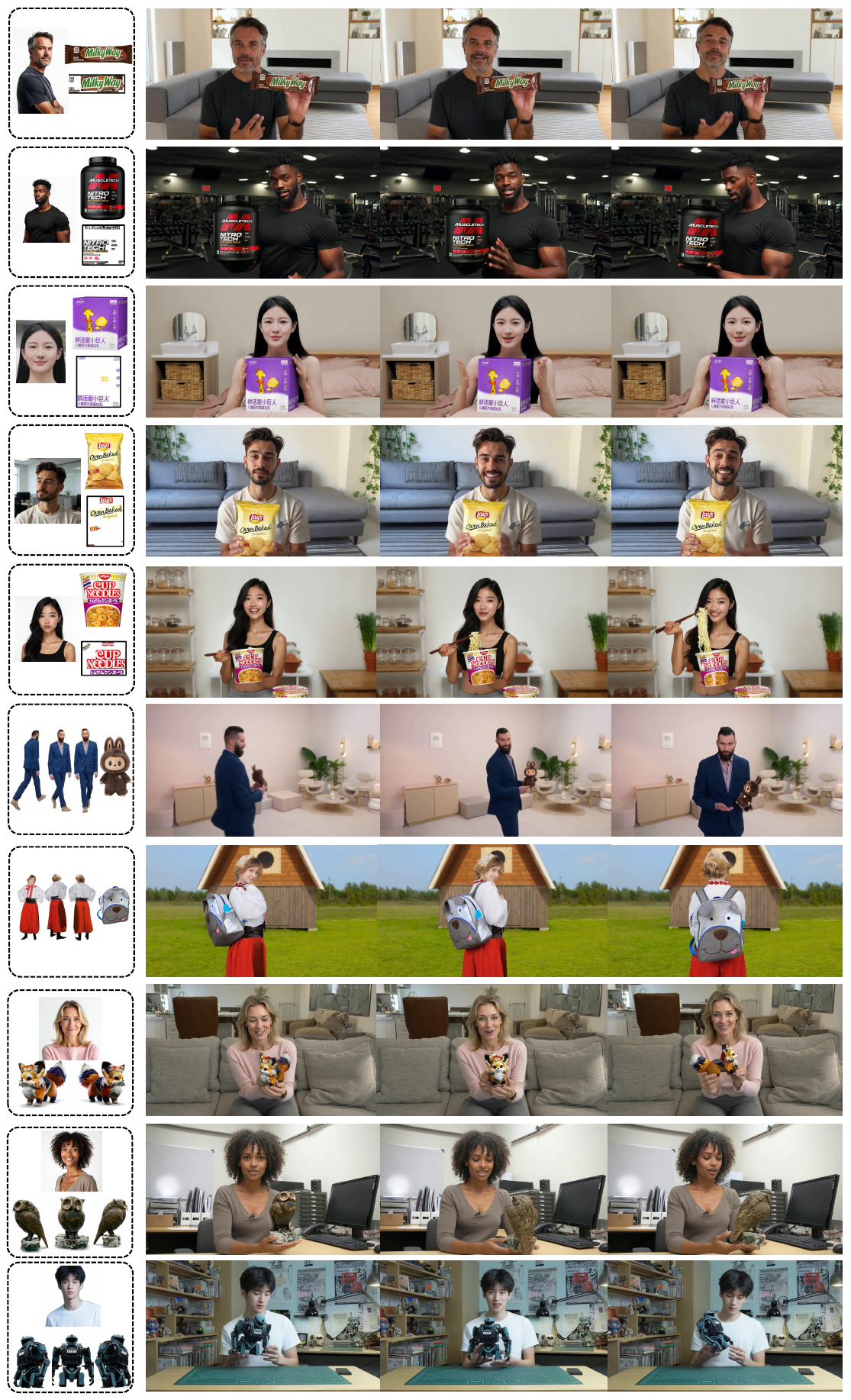}
    \caption{Additional visual results of HOMIE on intra-subject HOCVP scenarios, including OCR map and multi-view enhancements.}
    \label{fig:supp_more_results_2}
\end{center}
\end{figure*}

\noindent \textbf{More Qualitative Results.} This section provides additional visual results across a wider range of objects (Fig.~\ref{fig:supp_more_results_1} and Fig.\ref{fig:supp_more_results_2}). HOMIE successfully handles diverse human-object centric combinations, including but not limited to daily necessities, instruments, clothing, animals and furniture. Challenging HOCVP tasks that include abstract concepts as references are also included. Both inter- and intra-subject reference scenarios are included. 

\begin{figure*}[h]
\begin{center}
    \includegraphics[width=1.00\linewidth]{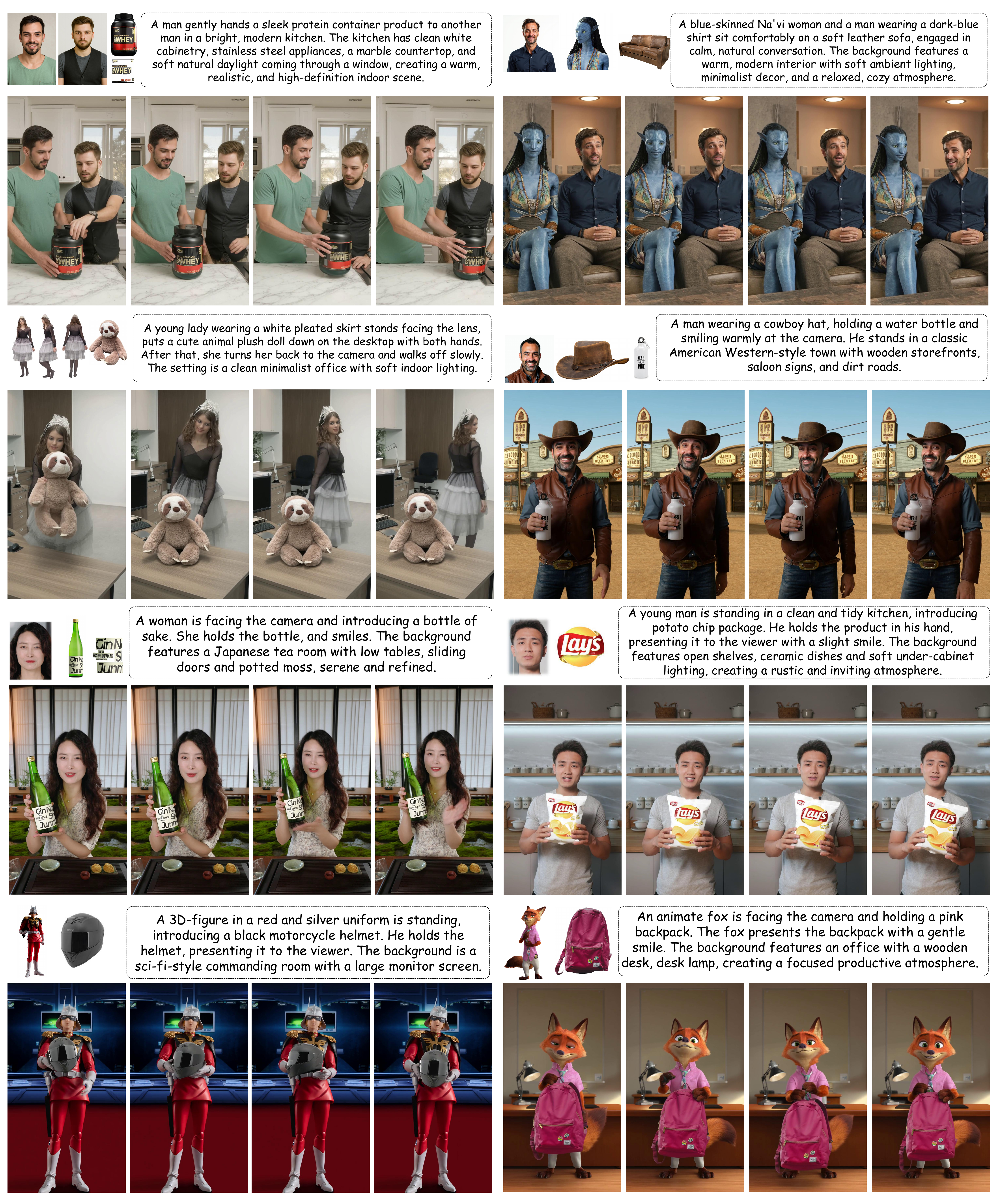}
    \caption{Demonstration of real application (e.g., product promotion)}
    \label{fig:supp_real_application}
\end{center}
\end{figure*}

\noindent \textbf{Practical Application.} We demonstrate HOMIE’s potential for real-world deployment in applications such as live-stream product promotion and creative video generation (Fig.~\ref{fig:supp_real_application}). With stable support for $1280 \times 720$ resolution, which is readily applicable to mobile devices, the model is highly suitable for practical use.

\textbf{All video samples are available on our anonymous demo webpage.} 

% ============ MAIN FIGURE HERE!!! ===============

\subsection{More Discussions}
\label{sec:more_discussion}
\subsubsection{Discussion of Abstract Concept Personalization}
We note that abstract concept references in video personalization are also explored by the concurrent work BrandFusion \cite{zhu2026brandfusion}. BrandFusion employs a multi-agent framework, combining MLLMs and T2V models, to naturally place a logo at a target position within a video. This supports our view that MLLM knowledge can help video models handle abstract concept personalization. However, the core contribution of HOMIE lies in its unified framework, which integrates MLLM reasoning through model design rather than relying on a multi-agent workflow.

\subsubsection{Ethical Concerns}
As noted in Sec.\ref{sec:supp_eval}, the human reference images utilized herein are sourced from public datasets or AIGC-generated, thereby circumventing human-subject risks. Nevertheless, given the generative nature of HOCVP, we acknowledge potential misuse risks, such as deepfakes. To mitigate these, future releases will enforce strict compliance guidelines to prevent malicious applications and ensure ethical deployment.

\subsubsection{Limitation}
\label{sec:limitation}
Built upon the Wan-T2V-14B series, HOMIE inevitably inherits the limitations of its foundational models. For instance, since the underlying Wan architecture is optimized for generating short clips of approximately 5 seconds, our framework currently shares this duration limit. Nevertheless, all competitive baselines evaluated herein are similarly constrained to this specific length. In future work, we aim to extend our methodology to long video generation, broadening its practical applications.